\documentclass[letterpaper]{article}

\usepackage[utf8]{inputenc} \usepackage[T1]{fontenc}    \usepackage[a4paper, total={6in, 8in}]{geometry}

\usepackage{amsmath,amssymb}
\usepackage{microtype}
\usepackage{graphicx}
\usepackage{subfigure}
\usepackage{booktabs}
\usepackage{hyperref}

\usepackage{subfigure}
\usepackage{tikz}
\usetikzlibrary{positioning}

\usepackage{pgfplots}
\usepgfplotslibrary{fillbetween}
\pgfplotsset{compat=1.14}

\usepackage{enumitem}
\setlist{itemsep=0pt,parsep=1pt,leftmargin=*}
\usepackage{amsthm,amsmath,amssymb}
\usepackage{url}
\usepackage{nicefrac}
\usepackage{mathtools}
  
\theoremstyle{definition}
\newtheorem{theorem}              {Theorem}

\newtheorem{problem}    [theorem] {Problem}
\newtheorem{example}    [theorem] {Example}

\newtheorem{assumption} [theorem] {Assumption}

\numberwithin{equation}{section}
\numberwithin{figure}{section}
\numberwithin{table}{section}

\newcommand{\bench}{Z}

\renewcommand{\Pr}{\mathbb{P}}

\title{Faking Fairness via Stealthily Biased Sampling}
\author{Kazuto Fukuchi~\thanks{The authors are listed in alphabetical order.}~\thanks{University of Tsukuba, Japan.}~\thanks{RIKEN AIP, Japan. \url{{kazuto.fukuchi, takanori.maehara}@riken.jp}} \and Satoshi Hara~\thanks{Osaka University, Japan. \url{satohara@ar.sanken.osaka-u.ac.jp}} \and Takanori Maehara~\footnotemark[3]}
\date{}

\begin{document}

\maketitle

\begin{abstract}
  Auditing fairness of decision-makers is now in high demand.
  To respond to this social demand, several fairness auditing tools have been developed.
The focus of this study is to raise an awareness of the risk of malicious decision-makers who fake fairness by abusing the auditing tools and thereby deceiving the social communities.
The question is whether such a fraud of the decision-maker is detectable so that the society can avoid the risk of fake fairness.
  In this study, we answer this question \emph{negatively}.
We specifically put our focus on a situation where the decision-maker publishes a benchmark dataset as the evidence of his/her fairness and attempts to deceive a person who uses an auditing tool that computes a fairness metric.
  To assess the (un)detectability of the fraud, we explicitly construct an algorithm, the stealthily biased sampling, that can deliberately construct an evil benchmark dataset via subsampling. 
  We show that the fraud made by the stealthily based sampling is indeed difficult to detect both theoretically and empirically.
  \end{abstract}
  
\section{Introduction}
\label{sec:introduction}

\paragraph{Background}

Machine learning models are being increasingly used in individuals' consequential decisions such as loan, insurance, and employment.
In such applications, the models are required to be \emph{fair} in the sense that their outputs should be irrelevant to the individuals' sensitive feature such as gender, race, and religion~\cite{pedreshi2008discrimination}.
Several efforts have been devoted to establishing mathematical formulation of fairness~\cite{dwork2012fairness,hardt2016equality,dwork2018individual} and to propose algorithms that meet the fairness criteria~\cite{bolukbasi2016man,feldman2015certifying,joseph2016fairness}.

With increasing attention to fairness, social communities now require to audit systems that incorporate machine learning algorithms to prevent unfair decisions.
For example, a 2014 White House Report~\cite{podesta2014big} mentioned ``[t]he increasing use of algorithms to make eligibility decisions must be carefully monitored for potential discriminatory outcomes for disadvantaged groups, even absent discriminatory intent''.
A similar statement also appeared in a 2016 White House Report~\cite{munoz2016big}.

To respond to the above social request, several fairness auditing tools have been developed~\cite{2016fairml,bellamy2018ai,2018aequitas}.
These tools help the decision-maker to investigate the fairness of their system by, e.g., computing several fairness metrics~\cite{2018aequitas}, measuring the significance of the system inputs~\cite{2016fairml}, and identifying minority groups with unfair treatments~\cite{bellamy2018ai}.
If the decision-maker found unfairness in their systems, she/he can then fix the systems by inspecting the causes of unfairness.

These auditing tools are also useful for promoting fairness of the decision-maker's system to the social communities.
For promoting fairness of the system, the decision-maker publishes the outputs of the auditing tools.
If the outputs suggest no unfairness in the system, the fact can be seen as an evidence of the system's fairness.
The decision-maker thus can appeal fairness of their system by publishing the fact to earn the trust of the social communities.

\paragraph{Risk of Fake Fairness}

The focus of this study is to raise awareness of the potential risk of malicious decision-makers who fake fairness.
If the decision-maker is malicious, he may control the auditing tools' results so that his system looks fair for the social communities even if the system is indeed unfair.
Such a risk is avoidable if the social communities can detect the decision-maker's fraud.
Therefore, the question is whether such a fraud is detectable.
In this study, we answer this question \emph{negatively}.
That is, the fraud is very difficult to detect in practice, which indicates that the society is now facing a potential risk of fake fairness.
In what follows, we refer to a person who attempts to detect the decision-maker's fraud as a \emph{detector}.

If the decision-maker only publishes the auditing tools' outputs, the detectability of the decision-maker's fraud is considerably low.
That is, the malicious decision-maker may modify the auditing tools' outputs arbitrary, whereas the detector has no way to certify whether or not the outputs are modified.
This concludes that the decision-maker who only publishes the auditing tools' outputs might be untrustable.

The decision-maker should publish more information about their system in addition to the auditing tools' outputs to acquire the social communities' trust.
However, because the system's information usually involves some confidential information, the decision-maker wants to prove fairness by publishing minimal information about their system.

In this study, we investigate a decision-maker who attempts to prove the fairness of their system by constructing a \emph{benchmark dataset}.
That is, the decision-maker publishes a subset of his dataset with his decisions as minimal information for proving the fairness of the system.
Given the benchmark dataset with the decisions, the detector can confirm the fairness of the system by using the auditing tools. In particular, we focus our attention on an auditing tool that computes a fairness metric. With this setup, we assess the detectability of the decision-maker's fraud. 

\paragraph{Biased Sampling Attack}

With the setup above, we consider a type of decision-maker's attacking algorithm, \emph{biased sampling attack}. 
In the biased sampling attack, an attacker has a dataset $D$ obtained from an underlying distribution $P$.
Here, the dataset $D$ involves the decisions made by the decision-makers' system, which are possibly unfair. 
The attacker deliberately selects a subset $\bench \subseteq D$ as the benchmark dataset so that the value of the fairness metric for $\bench$ is within a fair level. 
Then, the detector who employs an auditing tool that computes the fairness metric cannot detect unfairness of the decision-maker's system. 

The simplest method of the biased sampling attack might be \emph{case-control sampling}~\cite{mantel1959statistical}.
If the sensitive information is gender (man or woman) and the decision is binary (positive or negative), this method classifies the dataset into four classes: (man, positive), (woman, positive), (man, negative), and (woman, negative).
Then, it samples the desired numbers of points from the classes.
By controlling the number of points in each class appropriately, it produces a fair subset $\bench$.

Fortunately, the fraud of the case-control sampling could be detected as follows. 
The detector compares the distribution of the benchmark dataset $\bench$ with her prior knowledge (e.g., distributions of ages or zip-codes).
Then, because the case-control samples involve a bias from the original distribution, the detector may discover some unnatural thing, which indicates the decision-maker's fraud in the data-revealing process.

To hide the fraud, the malicious decision-maker will select fair subset $\bench$ whose distribution looks similar to that of $D$.
We refer to such a subset as \emph{stealthily biased subset} and the problem of sampling such a subset as \emph{stealthily biased sampling}.
Intuitively, the problem is formulated as follows. 
The mathematical formulation of the problem is given in Section~\ref{sec:algorithm}.
\begin{problem}[Stealthily biased sampling problem (informal)]
  \label{prob:stealth}
  Given a possibly unfair dataset $D$ obtained from an underlying distribution $P$, sample subset $\bench \subseteq D$ such that
  (i) $\bench$ is fair in terms of some fairness criteria, and 
  (ii) the distinguishing of the distribution of $\bench$ from $P$ is difficult.
\end{problem}

\paragraph{Our Contributions}

In this study, we develop an algorithm for the stealthily biased sampling problem and demonstrate its difficulty of detection.

First, we formulate the stealthily biased sampling problem as a \emph{Wasserstein distance minimization problem}. 
We show that this problem is reduced to the \emph{minimum-cost flow problem} and solved it in polynomial time. (Section~\ref{sec:algorithm})

Second, we show the difficulty of the detection of the proposed algorithm.
We introduce an ideal detector who can access the underlying distribution $P$ and compares the distribution of $\bench$ and $P$ by a statistical test.
The ideal detector has full information to perform the previously-mentioned fraud detection procedure, and any realistic detector cannot have such access.
Therefore, if the ideal detector cannot detect the fraud, we can conclude that any realistic detector either cannot detect the fraud.
We prove that the Wasserstein distance is an upper-bound of the \emph{advantage}, which is a distinguishability measure used in the cryptographic theory~\cite{goldreich2009foundations}, with respect to the Kolmogorov--Smirnov statistical test (KS test)~\cite{massey1951kolmogorov} (Theorem~\ref{thm:adv-ks-our}).
This means that the proposed algorithm is hard to detect even if the ideal detector uses the KS test.
(Section~\ref{sec:stealthiness})

Finally, through synthetic and real-world data experiments, we show that the decision-maker can indeed pretend to be fair by using the stealthily biased sampling.
Specifically, we demonstrate that the detector cannot detect the fraud of the decision-maker.
In the experiments, we investigate detectability against a detector who can access an independent observation from $P$ but cannot $P$. 
This detector is also ideal but more practical than the detector introduced in Section~\ref{sec:stealthiness}. 
The experimental results thus show more practical detectability than the theoretically analyzed one.
(Section~\ref{sec:experiments})

\section{Preliminaries}

\paragraph{Wasserstein Distance}

Let $V$ be a finite set, and $\mu, \nu \colon V \to \mathbb{R}_{\ge 0}$ be measures on $V$.  
A measure $\pi$ on $V \times V$ is a \emph{coupling measure} of $\mu$ and $\nu$ if $\mu_i = \sum_{j \in V} \pi_{ij}, \ \ \text{and} \ \ \nu_j = \sum_{i \in V} \pi_{ij}$, 
which is denoted by $\pi \in \Delta(\mu, \nu)$.
Let $(\mathcal{X}, d)$ be a metric space, i.e., $d \colon \mathcal{X} \times \mathcal{X} \to \mathbb{R}$ is positive definite, symmetric, and satisfies the triangle inequality.
Suppose that each $i \in V$ has feature $x_i \in \mathcal{X}$ on the metric space.
Then, the \emph{Wasserstein distance between $\mu$ and $\nu$}, denoted by $W(\mu, \nu)$, is defined by the optimal value of the following optimization problem~\cite{vaserstein1969markov}:
\begin{align}
  \text{min} \displaystyle \sum_{i, j \in V} d(x_i, x_j) \pi_{ij}, \ \text{s.t.} \ \ \pi \in \Delta(\mu, \nu).
\end{align}
The Wasserstein distance is computed in polynomial time by reducing to the minimum-cost flow problem or using the Sinkhorn iteration~\cite{peyre2017computational}.

\paragraph{Minimum-Cost Flow}

Let $\mathcal{G} = (\mathcal{V}, \mathcal{E})$ be a directed graph, where $\mathcal{V}$ is the vertices, $\mathcal{E}$ is the edges, $c \colon \mathcal{E} \to \mathbb{R}$ is the capacity, and $a \colon \mathcal{E} \to \mathbb{R}$ is the cost.
The minimum-cost flow problem is given by
\begin{align}
\begin{array}{ll}
\text{min} & \displaystyle \sum_{e \in \mathcal{E}} a(e) f(e) \\
  \text{s.t.} & 0 \le f(e) \le c(e), \  e \in \mathcal{E}, \\
  & \displaystyle \smashoperator{\sum_{e \in \delta^+(u)}} f(e) - \smashoperator{\sum_{e \in \delta^-(u)}} f(e)= \begin{cases} 
  0, & u \in \mathcal{V} \setminus \{s, t\}, \\
  d, & u = s, \\
  -d, & u = t,
  \end{cases} 
\end{array}
\end{align}
where $\delta^+(u) = \{ (u, v) \in \mathcal{E} \}$ and $\delta^-(v) = \{ (u, v) \in \mathcal{E} \}$ are the outgoing edges from $u$ and the incoming edges to $v$, respectively.
$d \ge 0$ is the required amount of the flow.
This problem is solvable in $\tilde O(\mathcal{E} \sqrt{\mathcal{V}})$ time in theory~\cite{lee2013path}, where $\tilde O$ suppresses $\log$ factors.
The practical evaluation of the minimum-cost flow algorithms are given in the study by \cite{kovacs2015minimum}.

\section{Algorithm for Stealthily Biased Sampling}
\label{sec:algorithm}

We formulate the stealthily biased sampling problem as a Wasserstein distance minimization problem.
The difficulty of detecting the stealthily biased sampling is studied in Section~\ref{sec:stealthiness}.
Here, we present a formulation for ``categorical biasing,'' which controls the number of points in each category.
Another biasing method for quantitative biasing is presented in Appendix~\ref{sec:general-biasing}.

\paragraph{Problem Formulation}

Let $\mathcal{X}$ be a metric space for the feature space and $\mathcal{Y}$ be a finite set representing the outcome of the decisions.
An entry of $x \in \mathcal{X}$ corresponds to a sensitive information; let $\mathcal{S}$ be a finite set representing the class of sensitive information, and let $s \colon \mathcal{X} \to \mathcal{S}$ be the mapping that extracts the sensitive information from the feature.

The dataset is given by $D = \{ (x_1, y_1), \dots, (x_N, y_N) \}$, where $x_i \in \mathcal{X}$ is the feature of the $i$-th point and $y_i \in \mathcal{Y}$ is the decision of the $i$-th point.
For simplicity, we write $i \in D$ for $(x_i, y_i) \in D$.

Let $\nu$ be the uniform measure on $D$, whose expected number of points is $K$, i.e., $\nu_i = \frac{K}{N}, (i \in D)$.
This is our reference distribution, i.e., if the decision-maker is not cheating, he will disclose subset $\bench \subseteq D$ following this distribution, i.e., $\Pr(i \in S) = \nu_i$, where $\Pr$ denotes the probability.

However, as the decision-maker wants to show that the output is fair, he constructs another distribution $\mu$.
Similar to the case-control sampling discussed in Section~\ref{sec:introduction}, we classify the dataset into bins $\mathcal{S} \times \mathcal{Y}$, and control the expected number of points sampled from each bin.
Let $k \colon \mathcal{S} \times \mathcal{Y} \to \mathbb{Z}$ be the number of points of the bins, where $K = \sum_{s \in \mathcal{S}, y \in \mathcal{Y}} k(s,y) \le |D|$.
Then, $\mu$ satisfies the requirement if 
\begin{align}
\label{eq:case-polytope}
 \sum_{\substack{(x_i, y_i) \in D : s(x_i) = s, y_i = y}} \mu_i = k(s,y), \ \ (s \in \mathcal{S}, y \in \mathcal{Y}).
\end{align}
We denote by $\mu \in P(k)$ if $\mu$ satisfies the above constraint.
Note that by choosing $k$ appropriately, we can show that $\bench$ is fair, thus meeting the first requirement in Problem~\ref{prob:stealth}.

To meet the second requirement in Problem~\ref{prob:stealth}, the decision-maker must determine distribution $\mu$ such that $\mu$ is indistinguishable from reference distribution $\nu$.
Here, we propose to measure the indistinguishability by using the Wasserstein distance.
Then, the stealthily biased sampling problem is mathematically formulated as follows.

\begin{problem}[Stealthily biased sampling problem (formal)]
\label{prob:stealth-formal}
$\text{min} \: W(\mu, \nu), \text{s.t.} \ \ \mu \in P(k).$
\end{problem}
By substituting the definition of the Wasserstein distance into Problem~\ref{prob:stealth-formal}, we obtain
\begin{align}
\label{eq:stealthilyLP}
  \text{min} \sum_{i, j \in D} d(x_i, x_j) \pi_{ij},  \text{s.t.} \ \ \pi \in \Delta(\mu, \nu), \mu \in P(k).
\end{align}
As the objective function is linear in $\pi$ and both $\Delta(\mu, \nu)$ and $P(k)$ are polytopes, Problem~\ref{eq:stealthilyLP} is a linear programming problem, hence is solved in a polynomial time~\cite{grotschel1981ellipsoid}.

\paragraph{Efficient Algorithm}

\begin{figure}
  \centering
\vspace{-24pt}
  \begin{tikzpicture}[scale=0.48]
  \tikzstyle{every node}=[font=\scriptsize]
  \node[draw,circle,label=north:$s$] at (0,0) (s) { };
  \node[draw,circle,label=north:$u_{sy}$] at (2,3) (u00) { };
  \node[draw,circle] at (2,1) (u01) { };
  \node[draw,circle] at (2,-1) (u10) { };
  \node[draw,circle] at (2,-3) (u11) { };
  \node[draw,circle,label=north:$l_i$] at (4,4) (l1) { };
  \node[draw,circle] at (4,3) (l2) { };
  \node[draw,circle] at (4,2) (l3) { };
  \node[draw,circle] at (4,1) (l4) { };
  \node[draw,circle] at (4,0) (l5) { };
  \node[draw,circle] at (4,-1) (l6) { };
  \node[draw,circle] at (4,-2) (l7) { };
  \node[draw,circle] at (4,-3) (l8) { };
  \node[draw,circle] at (4,-4) (l9) { };
  \node[draw,circle,label=north:$r_j$] at (8,4) (r1) { };
  \node[draw,circle] at (8,3) (r2) { };
  \node[draw,circle] at (8,2) (r3) { };
  \node[draw,circle] at (8,1) (r4) { };
  \node[draw,circle] at (8,0) (r5) { };
  \node[draw,circle] at (8,-1) (r6) { };
  \node[draw,circle] at (8,-2) (r7) { };
  \node[draw,circle] at (8,-3) (r8) { };
  \node[draw,circle] at (8,-4) (r9) { };
  \node[draw,circle,label=north:$t$] at (10,0) (t) { };
  \draw[->] (s)--(u00) node [midway,left] {$(k(s,y), 0)$};
  \foreach \u in {u01,u10,u11}
      \draw[->] (s)--(\u);
  \draw[->] (u00)--(l1) node [midway,above] {$(1, 0)$};
  \foreach \u / \v in {u00/l2,u01/l3,u01/l4,u10/l5,u10/l6,u11/l7,u11/l8,u11/l9}
      \draw[->] (\u)--(\v);
  \draw[->] (l1)--(r1) node [midway,above] {$(\infty, d(x_i, x_j))$};
  \foreach \u in {l1,l2,l3,l4,l5,l6,l7,l8,l9}
      \foreach \v in {r1,r2,r3,r4,r5,r6,r7,r8,r9}
          \draw[->] (\u) -- (\v);
  \draw[->] (r1)--(t) node [midway,right] {$(K/N, 0)$};
  \foreach \u in {r2,r3,r4,r5,r6,r7,r8,r9}
      \draw[->] (\u) -- (t);
  \end{tikzpicture}
  \caption{Flow network for biased sampling. $(c, a)$ on the edge if it has capacity $c$ and cost $a$.}
  \label{fig:flow}
\vspace{-14pt}
\end{figure}
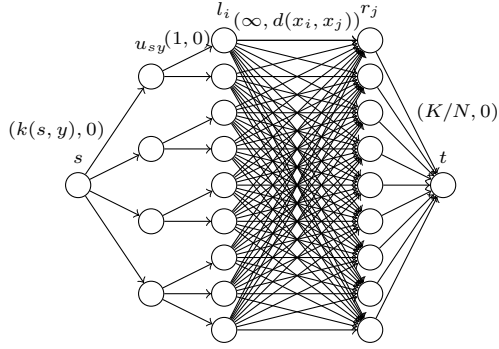

To establish an efficient algorithm for the stealthily biased sampling problem, we reduce the problem to a minimum-cost flow problem.

We construct the network $\mathcal{G} = (\mathcal{V}, \mathcal{E})$ in Figure~\ref{fig:flow} with capacity $c$ and cost $a$.
Vertices $\mathcal{V}$ consist of the following five classes:
(i) supersource $s$, (ii) case-controlling vertices $u_{sy}$ for all $s \in \mathcal{S}$ and $y \in \mathcal{Y}$, (iii) left vertices $l_i$ for all $i \in D$, (iv) right vertices $r_j$ for all $j \in D$, and (v) supersink $t$.
Edges $\mathcal{E}$ consist of the following four classes:
(i') $(s, u_{sy})$ for all $s \in \mathcal{S}$ and $y \in \mathcal{Y}$, whose cost is $0$ and capacity is $k(s,y)$, (ii') $(u_{sy}, l_i)$ for all $i \in D$ with $s(x_i) = s$ and $y_i = y$, whose cost is $0$ and capacity is one, (iii') $(l_i, r_j)$ for all $i \in D$ and $j \in D$, whose cost is $d(x_i, x_j)$ and capacity is $\infty$, and (iv') $(r_j, t)$ for all $j \in D$, whose cost is $0$ and capacity is $K/N$.

By setting the flow amount to $K$, the solution to the above instance gives the solution to the stealthily biased sampling problem, where $\pi_{ij}$ is the flow across edge $(l_i, r_j)$, and $\mu_i$ is the flow across edge $(u_{s(x_i) y_i}, l_i)$.
As $|\mathcal{V}| = O(|D|)$ and $|\mathcal{E}| = O(|D|^2)$, the problem is solvable in $\tilde O(|D|^{2.5})$ time~\cite{lee2013path}.

\section{Stealthiness of Sampling}
\label{sec:stealthiness}

We theoretically confirm that the stealthily biased sampling is difficult to detect.
Recall that the decision-maker's purpose is to make distribution $\mu$ indistinguishable from the uniform distribution $\nu$.
To measure the indistinguishability, we introduce \emph{advantage}, which is used in cryptographic theory~\cite{goldreich2009foundations}.

Let $\nu^K$ be $K$ product distribution of a sample drawn from the uniform probability distribution, and let $\mu^K$ be $K$ product distribution of a sample generated by our stealthily biased sampling algorithm. To define the advantage, let us consider the following game in which a detector attempts to distinguish $\mu^K$ and $\nu^K$:
(1). Flip an unbiased coin.
(2). If a head outcome is achieved, the decision-maker reveals $D \sim \mu^K$ to the detector; otherwise, the decision-maker reveals $D \sim \nu^K$ to the detector.
(3). The detector estimates the side of the flipped coin. 
If the probability that the detector estimates the outcome of the unbiased coin correctly is near $1/2$, the detector cannot distinguish whether the obtained samples are biased. 

Let $H$ be a random variable such that $\Pr(H = 1) = \Pr(H = 0) = 1/2$, which represents the flipped unbiased coin.
The detector's estimation algorithm is a mapping $\Phi$ from $D$ to $\{0, 1\}$, where the output is $1$ if the detector expects that the samples are drawn from $\nu^K$; otherwise, the output is $0$.
The probability that the detector detects bias correctly is obtained as $\Pr(\Phi(D) = H)$, where the randomness comes from flipped coin $H$ and dataset $D$.
Then, the advantage is defined as follows:
\begin{align}
\mathrm{Adv}(\Phi;\mu^K,\nu^K) = \left|\Pr(\Phi(D) = H) - \tfrac{1}{2}\right|.
\end{align}
A smaller $\mathrm{Adv}$ value implies that biased distribution $\mu^K$ is more difficult to distinguish from $\nu^K$ against a detector with the test algorithm $\Phi$.

\paragraph{Stealthiness against Kolmogorov--Smirnov Test}

To assess the difficulty of detecting the stealthily biased sampling, we consider an ideal detector who can access the underlying distribution $\nu$.
Here, we analyze the advantage when the ideal detector who uses the KS test.

The KS test is a goodness-of-fit test for real-valued samples. Let $F_\nu$ be the cumulative distribution function of distribution $\nu$, and let $F_K$ be the cumulative distribution function of the empirical measure of the obtained samples.
Then, the KS statistic is defined as $\mathrm{KS}(D;\nu) = \sup_{x}|F_K(x) - F_\nu(x)|$.
The KS test is rejected if $\mathrm{KS}(D;\nu)$ is larger than an appropriate threshold. 

Let us consider the detector's algorithms based on the KS statistic.
We formally define a detector's algorithm that returns $1$ if the KS statistic is larger than threshold $\tau$ as $\Phi_{\mathrm{KS},\tau}(D) = \mathbb{I}(\mathrm{KS}(D;\nu) > \tau)$, 
where $\mathbb{I}$ is the indicator function.

We analyze the advantage against $\Phi_{\mathrm{KS},\tau}$ under a flatness assumption on sample distribution $\nu$.
For $x \in \mathcal{X}$, let $B_\epsilon(x)$ be the $\epsilon$-ball centered at $x$.
Then, the flatness assumption is defined as follows:
\begin{assumption}\label{asm:flatness}
There exist constants $s, C > 0$ such that for any $\epsilon > 0$, $\sup_{x \in \mathcal{X}}\nu(B_\epsilon(x)) \le \left(\nicefrac{\epsilon}{C}\right)^s$.
\end{assumption}
Many natural distributions on a real line satisfy Assumption~\ref{asm:flatness}.
For example, the one-dimensional normal distribution satisfies Assumption \ref{asm:flatness} with $s = 1$ and $C=\sqrt{\nicefrac{2}{\pi}}$.

Under the flatness assumption on $\nu$, we reveal an upper bound on the advantage against the KS test in the categorical biasing setting. Let $M$ be the number of pair types of decision and sensitive attribute.
Let $\kappa$ and $\kappa'$ be the distribution over pairs of decision and sensitive attribute on the sample distribution and biased distribution. Then, we reveal the following theorem.
\begin{theorem}\label{thm:adv-ks-our}
Let $W(\mu^K, \nu^K)$ be the Wasserstein distance equipped with the distance $d(D,D') = \min_{i=1,...,K}d(x_i,x'_i)$ for $D=\{x_1,...,x_K\}$ and $D'=\{x'_1,...,x'_K\}$. Under Assumption \ref{asm:flatness}, for threshold $\tau \ge (C/K)^{1/s}/2$, we have   
\begin{align}
   \mathrm{Adv}(\Phi_{\mathrm{KS},\tau};\mu^K,\nu^K)     \le  \nicefrac{K^{1/s}W(\mu^K, \nu^K)}{C^{1/s}} + 4K!\left(\nicefrac{(1 + \mathrm{TV}(\kappa, \kappa'))}{K}\right)^K, \label{eq:adv-ks-our}
\end{align}
where $s$ and $C$ are the constants from Assumption \ref{asm:flatness}, $\mathrm{TV}(\kappa, \kappa') = \sum_i^M|\kappa_i - \kappa'_i|/2$ is the total variation distance.
\end{theorem}
The proof of this theorem can be found in the supplementary material.
Since $1 + \mathrm{TV}(\kappa, \kappa') < e$ and $K! \sim (K/e)^K$, the second term in \eqref{eq:adv-ks-our} is $o(1)$ and is dominated by the first term.
Because the stealthily biased sampling minimizes the Wasserstein distance (i.e., the first term of \eqref{eq:adv-ks-our}), it also minimizes the upper-bound of the advantage.
This implies that the stealthily biased sampling is difficult to detect for the ideal detector.
Consequently, for any realistic detector who has less information than the ideal one, it is even more difficult to detect the stealthily biased sampling.

\section{Experiments}
\label{sec:experiments}

In this section, we show that the stealthily biased sampling is indeed difficult to detect, through experiments on synthetic data and two real-world data (COMPAS and Adult).\footnote{The codes can be found at \url{https://github.com/sato9hara/stealthily-biased-sampling}}
In the experiments, we adopted the \emph{demographic parity} (DP)~\cite{calders2009building} as the fairness metric for auditing.
Here, let $s \in \{0, 1\}$ be a sensitive feature and $y \in \{0, 1\}$ be a decision.
The DP is then defined as $\mathrm{DP} = \left|\Pr (y=1 \mid s=1) - \Pr (y=1 \mid s=0 )\right|$.
A large DP indicates that the decision is unfair because the decision-maker favors providing positive decisions to one group over the other group.

\paragraph{Summary of the Results}
Before moving to each experiment, we summarize the main results here.
In the experiments, we investigated detectability of the decision-makers' fraud against an ideal detector who can access an independent observation $D'$ from the underlying distribution $P$.
In all the experiments, we verified the following three points.
\begin{enumerate}[label={\textbf{R\arabic*.}}]
  \item Both the stealthily biased and case-control sampling could reduce the DP of the sampled set $\bench$.
  \item The stealthily biased sampling was more resistant against the detector's fraud detection compared to the case-control sampling.
  Specifically, the stealthily biased sampling marked low scores of the fraud detection criteria for a wide range of the experimental settings.
  \item In all the experiments, the decision-makers successfully pretended to be fair. They could select a subset $\bench$ with small DPs and small fraud detection criteria.
\end{enumerate}

\paragraph{Implementation}
We used Python~3 for data processing.
In all the experiments, we used the squared Euclidean distance $d(x_i, x_j) = \|x_i - x_j\|^2$ as the metric in Wasserstein distance.
To solve the minimum-cost flow problem~\eqref{eq:stealthilyLP}, we used the network simplex method implemented in LEMON Graph Library.\footnote{https://lemon.cs.elte.hu/trac/lemon}
With LEMON, the problem could be solved in a few seconds for the datasets with the size $N$ up to a few thousand.
For the Adult dataset, we used a bootstrap-type estimator to improve the computational scalability (see Appendix~\ref{app:bootstrap} for the detail).

\subsection{Synthetic Example}
\label{sec:synthetic}

\begin{example}[Loan check]
  Consider a decision-maker who decides to lend money ($y=1$) or not ($y=0$) based on the applicants sensitive feature $s \in \{0, 1\}$ (e.g., gender) and a $d$-dimensional feature vector $x \in [0, 1]^d$, where first feature $x_1$ is an income.
  Here, we model the criteria of the decision-maker as
  \begin{align}
      y = \mathbb{I}(x_1 + b s > 0.5) ,
      \label{eq:simple_decision}
  \end{align}
  where $b \ge 0$ is a constant.
  Note that this decision-maker is unfair if $b \neq 0$.
\end{example}

To pretend to be a fair, for a set of individual's feature, sensitive feature, and the decision $D = \{(x_i, s_i, y_i)\}_{i=1}^N$, the decision-maker selects subset $\bench \subseteq D$ as evidence that the decisions are fair.
We solve this problem by using both the stealthily biased and case-control sampling.

\medskip
\noindent
\textbf{Data \:}
We set the underlying data distribution $P$ as follows.
We sampled sensitive feature $s$ with $\Pr ( s=1 ) = 0.5$, and sampled feature vector $x$ in a uniformly random manner over $[0, 1]^d$ with $d=1$.\footnote{Results for higher dimensional settings were almost the same as $d=1$. See Appendix~\ref{app:synthetic_deterministic}.}
Decision $y$ is made by following the criteria~\eqref{eq:simple_decision}.
We sampled dataset $D$ with $N=1,000$ observations from the underlying distribution $P$.
We set the parameters in the criteria~\eqref{eq:simple_decision} to be $b=0.2$.
Thus, the DP of the decision-maker is $0.2$.

\medskip
\noindent
\textbf{Attacker \:}
To reduce the DP through sampling, the sampled set needs to satisfy $\Pr ( y=1 \mid s=1 ) \approx \Pr ( y=1 \mid s=0 ) \approx \alpha$ for a predetermined ratio of positive decisions $\alpha \in [0, 1]$.
The expected number of sampling in each bin $(s, y) \in \{0, 1\} \times \{0, 1\}$ is then determined by $k(s,y) = \lceil 0.5 K  \alpha^y (1 - \alpha)^{1-y} \rceil$ (recall that $\Pr ( s ) =0.5, \forall s \in \{0, 1\}$).

\medskip
\noindent
\textbf{Detector \:}
As a detector, we adopted the Kolmogorov--Smirnov two-sample test.
The detector has an independent observation $D' = \{ (x'_j, s'_j) \}_{j=1}^{200}$ as a referential dataset sampled from underlying distribution $P$.
Here, we note that the detector has no access to decision $y$ for $D'$ because the decision criteria~\eqref{eq:simple_decision} is not disclosed.
Given $\bench$, the detector applies the Kolmogorov--Smirnov two-sample test to detect whether the distribution of $S$ is different from that of referential set $D'$.
Here, we consider the strongest detector: we assume that she knows that only income $x_1$ is used in $x$ for the decision.
We denote the distribution of income $x_1$ in $\bench$ and $D'$ by $\Pr_S ( x_1 )$ and $\Pr_{D'} ( x_1 )$, respectively.
The detector can then use the Kolmogorov--Smirnov two-sample test\footnote{In practice, the detector does not know that $x_1$ is a key feature. Thus, the detector needs to use the two-sample test for multi-dimensional data. However, in our preliminary experiments, we found that multi-dimensional tests have very low detection powers. 
Therefore, we used an advantageous setting for the detector.} in three ways: (i) test $\Pr_S ( x_1 ) = \Pr_{D'} ( x_1 )$, (ii) test $\Pr_S ( x_1 \mid s=1 ) = \Pr_{D'} (x_1 \mid s=1 )$, and (iii) test $\Pr_S ( x_1 \mid s=0 ) = \Pr_{D'} ( x_1 \mid s=0 )$.
In the experiment, we set the significance level of the test to be $0.05$.

\begin{figure*}[t]
\centering
\subfigure[Demographic parity]{
\centering
\begin{tikzpicture}
\begin{axis}[
scale=0.8,
width=0.45\linewidth,
height=100pt,
xmin=0.4,
xmax=0.8,
ymin=0,
yticklabel style={
    /pgf/number format/fixed,
    /pgf/number format/precision=2
},
xlabel={Ratio of positive decisions $\alpha$ in sampling},
ylabel={Average DP},
legend style={at={(0.5,1.1)},anchor=south,legend columns=-1, font=\footnotesize}
]

\addplot[name path=case_upper,draw=none,forget plot] table[x=a,y expr=\thisrow{Case-control}+\thisrow{Case-control-std}, col sep=comma] {./csv/parity_deterministic_d001_q20.csv};
\addplot[name path=case_lower,draw=none,forget plot] table[x=a,y expr=\thisrow{Case-control}-\thisrow{Case-control-std}, col sep=comma] {./csv/parity_deterministic_d001_q20.csv};
\addplot [fill=red!30,opacity=0.5,forget plot] fill between[of=case_upper and case_lower];

\addplot[name path=proposed_upper,draw=none,forget plot] table[x=a,y expr=\thisrow{Proposed}+\thisrow{Proposed-std}, col sep=comma] {./csv/parity_deterministic_d001_q20.csv};
\addplot[name path=proposed_lower,draw=none,forget plot] table[x=a,y expr=\thisrow{Proposed}-\thisrow{Proposed-std}, col sep=comma] {./csv/parity_deterministic_d001_q20.csv};
\addplot [fill=blue!30,opacity=0.5,forget plot] fill between[of=proposed_upper and proposed_lower];

\addplot[ultra thick, red, dashed] table[x=a,y=Case-control, col sep=comma]{./csv/parity_deterministic_d001_q20.csv};
\addplot[ultra thick, blue] table[x=a,y=Proposed, col sep=comma]{./csv/parity_deterministic_d001_q20.csv};

\legend{Case-control, Stealth}

\end{axis}
\end{tikzpicture}
\label{fig:synthetic_deterministic_parity}
}
\subfigure[Ratio of rejection in test $\Pr(x)$]{
\centering
\begin{tikzpicture}
\begin{axis}[
scale=0.8,
width=0.45\linewidth,
height=100pt,
xmin=0.4,
xmax=0.8,
ymin=0,
xlabel={Ratio of positive decisions $\alpha$ in sampling},
ylabel style={align=center},
ylabel={Ratio of \\ rejection},
legend style={at={(0.47,0.97)},anchor=north,font=\footnotesize}
]

\addplot[name path=case_upper,draw=none,forget plot] table[x=a,y expr=\thisrow{Case-control}+\thisrow{Case-control-conf}, col sep=comma] {./csv/reject_deterministic_d001_q20_test0.csv};
\addplot[name path=case_lower,draw=none,forget plot] table[x=a,y expr=\thisrow{Case-control}-\thisrow{Case-control-conf}, col sep=comma] {./csv/reject_deterministic_d001_q20_test0.csv};
\addplot [fill=red!30,opacity=0.5,forget plot] fill between[of=case_upper and case_lower];

\addplot[name path=proposed_upper,draw=none,forget plot] table[x=a,y expr=\thisrow{Proposed}+\thisrow{Proposed-conf}, col sep=comma] {./csv/reject_deterministic_d001_q20_test0.csv};
\addplot[name path=proposed_lower,draw=none,forget plot] table[x=a,y expr=\thisrow{Proposed}-\thisrow{Proposed-conf}, col sep=comma] {./csv/reject_deterministic_d001_q20_test0.csv};
\addplot [fill=blue!30,opacity=0.5,forget plot] fill between[of=proposed_upper and proposed_lower];

\addplot[ultra thick, red, dashed] table[x=a,y=Case-control, col sep=comma]{./csv/reject_deterministic_d001_q20_test0.csv};
\addplot[ultra thick, blue] table[x=a,y=Proposed, col sep=comma]{./csv/reject_deterministic_d001_q20_test0.csv};
\addplot[ultra thick, black, dotted, domain=0:1] {0.05};

\end{axis}
\end{tikzpicture}
\label{fig:synthetic_deterministic_reject_x}
}

\subfigure[Ratio of rejection in test $\Pr(x|s=1)$]{
\centering
\begin{tikzpicture}
\begin{axis}[
scale=0.8,
width=0.45\linewidth,
height=100pt,
xmin=0.4,
xmax=0.8,
ymin=0,
xlabel={Ratio of positive decisions $\alpha$ in sampling},
ylabel style={align=center},
ylabel={Ratio of \\ rejection},
legend pos=north east
]

\addplot[name path=case_upper,draw=none,forget plot] table[x=a,y expr=\thisrow{Case-control}+\thisrow{Case-control-conf}, col sep=comma] {./csv/reject_deterministic_d001_q20_test1.csv};
\addplot[name path=case_lower,draw=none,forget plot] table[x=a,y expr=\thisrow{Case-control}-\thisrow{Case-control-conf}, col sep=comma] {./csv/reject_deterministic_d001_q20_test1.csv};
\addplot [fill=red!30,opacity=0.5,forget plot] fill between[of=case_upper and case_lower];

\addplot[name path=proposed_upper,draw=none,forget plot] table[x=a,y expr=\thisrow{Proposed}+\thisrow{Proposed-conf}, col sep=comma] {./csv/reject_deterministic_d001_q20_test1.csv};
\addplot[name path=proposed_lower,draw=none,forget plot] table[x=a,y expr=\thisrow{Proposed}-\thisrow{Proposed-conf}, col sep=comma] {./csv/reject_deterministic_d001_q20_test1.csv};
\addplot [fill=blue!30,opacity=0.5,forget plot] fill between[of=proposed_upper and proposed_lower];

\addplot[ultra thick, red, dashed] table[x=a,y=Case-control, col sep=comma]{./csv/reject_deterministic_d001_q20_test1.csv};
\addplot[ultra thick, blue] table[x=a,y=Proposed, col sep=comma]{./csv/reject_deterministic_d001_q20_test1.csv};
\addplot[ultra thick, black, dotted, domain=0:1] {0.05};

\end{axis}
\end{tikzpicture}
\label{fig:synthetic_deterministic_reject_x_s1}
}
\subfigure[Ratio of rejection in test $\Pr(x|s=0)$]{
\centering
\begin{tikzpicture}
\begin{axis}[
scale=0.8,
width=0.45\linewidth,
height=100pt,
xmin=0.4,
xmax=0.8,
ymin=0,
xlabel={Ratio of positive decisions $\alpha$ in sampling},
ylabel style={align=center},
ylabel={Ratio of \\ rejection},
legend pos=north west
]

\addplot[name path=case_upper,draw=none,forget plot] table[x=a,y expr=\thisrow{Case-control}+\thisrow{Case-control-conf}, col sep=comma] {./csv/reject_deterministic_d001_q20_test2.csv};
\addplot[name path=case_lower,draw=none,forget plot] table[x=a,y expr=\thisrow{Case-control}-\thisrow{Case-control-conf}, col sep=comma] {./csv/reject_deterministic_d001_q20_test2.csv};
\addplot [fill=red!30,opacity=0.5,forget plot] fill between[of=case_upper and case_lower];

\addplot[name path=proposed_upper,draw=none,forget plot] table[x=a,y expr=\thisrow{Proposed}+\thisrow{Proposed-conf}, col sep=comma] {./csv/reject_deterministic_d001_q20_test2.csv};
\addplot[name path=proposed_lower,draw=none,forget plot] table[x=a,y expr=\thisrow{Proposed}-\thisrow{Proposed-conf}, col sep=comma] {./csv/reject_deterministic_d001_q20_test2.csv};
\addplot [fill=blue!30,opacity=0.5,forget plot] fill between[of=proposed_upper and proposed_lower];

\addplot[ultra thick, red, dashed] table[x=a,y=Case-control, col sep=comma]{./csv/reject_deterministic_d001_q20_test2.csv};
\addplot[ultra thick, blue] table[x=a,y=Proposed, col sep=comma]{./csv/reject_deterministic_d001_q20_test2.csv};
\addplot[ultra thick, black, dotted, domain=0:1] {0.05};

\end{axis}
\end{tikzpicture}
\label{fig:synthetic_deterministic_reject_x_s0}
}
 \caption{
Results for the decision-maker with $a=0.2$. The shaded regions in (a) denotes the average DP $\pm$ std. The shaded regions in (b)--(d) denote 95\% confidence intervals. The dotted line in (b)--(d) denotes the significance level $0.05$.
}
\label{fig:synthetic_deterministic}
\end{figure*}
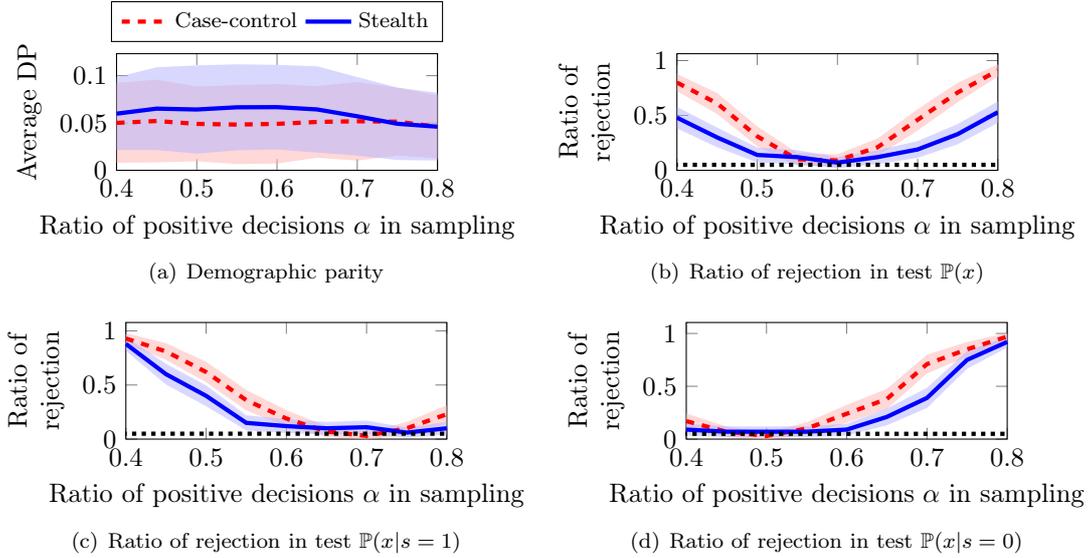

\medskip
\noindent
\textbf{Result \:}
We selected a subset $\bench \subseteq D$ with size $|\bench|=200$ using both the stealthily biased and case-control sampling.
We repeated the experiment 100 times, and summarized the results in \figurename~\ref{fig:synthetic_deterministic}, for several different ratios of positive decisions $\alpha$.
As we summarized earlier, three key observations R1, R2, and R3 can be found in the figures.

\medskip
\noindent
\textbf{R1.}
\figurename~\ref{fig:synthetic_deterministic_parity} shows that both the stealthily biased and case-control sampling successfully reduced DP to less than $0.1$ through sampling the subset $\bench$.
We note that no significant differences were observed in DPs between the two sampling methods.

\medskip
\noindent
\textbf{R2.}
Figures~\ref{fig:synthetic_deterministic_reject_x}, \ref{fig:synthetic_deterministic_reject_x_s1}, and \ref{fig:synthetic_deterministic_reject_x_s0} show that the stealthily biased sampling was more resistant to the Kolmogorov--Smirnov test, compared to the case-control sampling.
Specifically, the stealthily biased sampling attained a small rejection rate in a wide range of $\alpha$ in the sampling process.

\medskip
\noindent
\textbf{R3.}
By using the stealthily biased sampling, the decision-maker successfully pretended to be fair.
By setting $\alpha$ in the sampling to be $0.6$, none of the tests could confidently reject that disclosed dataset $\bench$ is different from the referential dataset $D'$.
For $\alpha = 0.6$, the rejection rates of all the three tests were kept around 0.05, which is exactly the same as the significance level.
These results indicate that the detector cannot detect the fraud made by the stealthily biased sampling: the DP of $\bench$ is small, and its distribution is sufficiently natural so that the statistical test cannot reject it.
The case-control sampling showed higher rejection rates in tests of $\Pr (x \mid s=1 )$ and $\Pr ( x \mid s=0 )$, and thus was outperformed by the stealthily biased sampling.

Lastly, we note that the stochastic decision-maker can be far more evil than the deterministic decision-maker considered in this section.
See Appendix~\ref{app:synthetic_stochastic} for the detail.

\subsection{Real-World Data: COMPAS}
\label{sec:compas}

For the first real-world data experiment, we focus on the COMPAS dataset~\cite{angwin2016machine}.\footnote{\url{https://github.com/propublica/compas-analysis}}
The COMPAS dataset contains several defendant's records obtained from the Broward County Sheriff’s Office in Florida.
Each defendant is scored his or her risk of recidivism using a software called COMPAS.
ProPublica~\cite{angwin2016machine} revealed that the COMPAS risk score is discriminative: it tends to score white defendants with low scores while scoring black defendants with high scores.

Because Florida had strong open-records laws, the entire COMPAS dataset was made public, and the bias in the COMPAS risk score was revealed.
Here, we consider a virtual scenario that the decision-maker was aware of the bias in the risk score, and he wants to pretend to be fair by hiding the bias.
To attain this goal, the decision-maker discloses a subset of the COMPAS dataset as evidence that the COMPAS risk score is fair.

\medskip
\noindent
\textbf{Data \:}
We used the same data preprocessing following the analysis of ProPublica~\cite{angwin2016machine}, which results in eight features $x \in \mathbb{R}^8$ of each defendant, with race as sensitive attribute $s \in \{0 (\text{black}), 1 (\text{white})\}$, and the decision $y \in \{0 (\text{low-risk}), 1 (\text{middle/high-risk})\}$.
The preprocessed data includes $5,278$ records, which we randomly held out $1,278$ records as the referential dataset $D'$ for the detector.
From the remaining $4,000$ records $D$, we sampled $2,000$ records as $\bench$ using both the stealthily biased and case-control sampling.
To reduce the DP in the sampling, we required the sampled set to satisfy $\Pr ( y=1 \mid s=1 ) \approx \Pr ( y=1 \mid s=0 ) \approx \alpha$ for some $\alpha \in [0, 1]$.

\medskip
\noindent
\textbf{Detector \:}
The detector tries to detect the bias in the disclosed dataset $\bench$ by comparing its distribution with the referential dataset $D'$.
In the experiment, we adopted the Wasserstein distance (WD) as the detector's detection criteria.\footnote{In COMPAS and Adult experiments, we did not adopt the multi-dimensional two-sample tests because they were too weak.}
If the WD between $\bench$ and $D'$ is sufficiently large, the detector can detect the bias in $\bench$, and thus the fraud of the decision-maker is revealed.

\medskip
\noindent
\textbf{Result \:}
We repeated the experiment 100 times by randomly changing the data splitting, and summarized the results in \figurename~\ref{fig:compas}.\footnote{Here, we measured the WD on $\Pr(x)$. The WD on $\Pr(x \mid s=1)$ and $\Pr(x \mid s=0)$ can be found in Appendix~\ref{app:compas_res}.}
As the baseline without any biased sampling, we computed DP and the WD for randomly sampled records from $D$, which are denoted as \emph{Baseline} in the figures.
The figures show the clear success of the stealthily biased sampling, as we summarized in R1, R2, and R3.
In \figurename~\ref{fig:compas}(a), with the stealthily biased sampling, the DPs of $\bench$ have reduced significantly (R1).
In Figures~\ref{fig:compas}(b), the WDs between $\bench$ and $D'$ were sufficiently small for $\alpha = 0.6$ so that they are completely indistinguishable from the baselines (R3).
The case-control sampling had higher WDs, and it was thus easier for the detector to detect (R2).

\begin{figure*}[t]
\centering
\subfigure[Demographic parity]{
\centering
\begin{tikzpicture}
\begin{axis}[
scale=0.8,
width=0.45\linewidth,
height=100pt,
xmin=0.25,
xmax=0.75,
ymin=0,
yticklabel style={
    /pgf/number format/fixed,
    /pgf/number format/precision=2
},
xlabel={Ratio of positive decisions $\alpha$ in sampling},
ylabel={Average DP},
legend style={at={(0.5,1.1)},anchor=south,legend columns=-1, font=\footnotesize}
]

\addplot[name path=original_upper,draw=none,forget plot] table[x=ypos,y expr=\thisrow{No-Sampling}+\thisrow{Np-Sampling-std}, col sep=comma] {./csv/compas_result_parity.csv};
\addplot[name path=original_lower,draw=none,forget plot] table[x=ypos,y expr=\thisrow{No-Sampling}-\thisrow{Np-Sampling-std}, col sep=comma] {./csv/compas_result_parity.csv};
\addplot [fill=black!30,opacity=0.5,forget plot] fill between[of=original_upper and original_lower];

\addplot[name path=case_upper,draw=none,forget plot] table[x=ypos,y expr=\thisrow{Case-control}+\thisrow{Case-control-std}, col sep=comma] {./csv/compas_result_parity.csv};
\addplot[name path=case_lower,draw=none,forget plot] table[x=ypos,y expr=\thisrow{Case-control}-\thisrow{Case-control-std}, col sep=comma] {./csv/compas_result_parity.csv};
\addplot [fill=red!30,opacity=0.5,forget plot] fill between[of=case_upper and case_lower];

\addplot[name path=proposed_upper,draw=none,forget plot] table[x=ypos,y expr=\thisrow{Proposed}+\thisrow{Proposed-std}, col sep=comma] {./csv/compas_result_parity.csv};
\addplot[name path=proposed_lower,draw=none,forget plot] table[x=ypos,y expr=\thisrow{Proposed}-\thisrow{Proposed-std}, col sep=comma] {./csv/compas_result_parity.csv};
\addplot [fill=blue!30,opacity=0.5,forget plot] fill between[of=proposed_upper and proposed_lower];

\addplot[ultra thick, black, dotted] table[x=ypos,y=No-Sampling, col sep=comma]{./csv/compas_result_parity.csv};
\addplot[ultra thick, red, dashed] table[x=ypos,y=Case-control, col sep=comma]{./csv/compas_result_parity.csv};
\addplot[ultra thick, blue] table[x=ypos,y=Proposed, col sep=comma]{./csv/compas_result_parity.csv};

\legend{Baseline, Case-control, Stealth}

\end{axis}
\end{tikzpicture}
\label{fig:compas_parity}
}
\hspace{12pt}
\subfigure[Wasserstein distance in $\Pr(x)$]{
\centering
\begin{tikzpicture}
\begin{axis}[
scale=0.8,
width=0.45\linewidth,
height=100pt,
xmin=0.25,
xmax=0.75,
xlabel={Ratio of positive decisions $\alpha$ in sampling},
ylabel={Average WD},
legend style={at={(0.6,0.97)},anchor=north,font=\footnotesize}
]

\addplot[name path=original_upper,draw=none,forget plot] table[x=ypos,y expr=\thisrow{No-Sampling}+\thisrow{Np-Sampling-std}, col sep=comma] {./csv/compas_result_distance_test00.csv};
\addplot[name path=original_lower,draw=none,forget plot] table[x=ypos,y expr=\thisrow{No-Sampling}-\thisrow{Np-Sampling-std}, col sep=comma] {./csv/compas_result_distance_test00.csv};
\addplot [fill=black!30,opacity=0.5,forget plot] fill between[of=original_upper and original_lower];

\addplot[name path=case_upper,draw=none,forget plot] table[x=ypos,y expr=\thisrow{Case-control}+\thisrow{Case-control-std}, col sep=comma] {./csv/compas_result_distance_test00.csv};
\addplot[name path=case_lower,draw=none,forget plot] table[x=ypos,y expr=\thisrow{Case-control}-\thisrow{Case-control-std}, col sep=comma] {./csv/compas_result_distance_test00.csv};
\addplot [fill=red!30,opacity=0.5,forget plot] fill between[of=case_upper and case_lower];

\addplot[name path=proposed_upper,draw=none,forget plot] table[x=ypos,y expr=\thisrow{Proposed}+\thisrow{Proposed-std}, col sep=comma] {./csv/compas_result_distance_test00.csv};
\addplot[name path=proposed_lower,draw=none,forget plot] table[x=ypos,y expr=\thisrow{Proposed}-\thisrow{Proposed-std}, col sep=comma] {./csv/compas_result_distance_test00.csv};
\addplot [fill=blue!30,opacity=0.5,forget plot] fill between[of=proposed_upper and proposed_lower];

\addplot[ultra thick, black, dotted] table[x=ypos,y=No-Sampling, col sep=comma]{./csv/compas_result_distance_test00.csv};
\addplot[ultra thick, red, dashed] table[x=ypos,y=Case-control, col sep=comma]{./csv/compas_result_distance_test00.csv};
\addplot[ultra thick, blue] table[x=ypos,y=Proposed, col sep=comma]{./csv/compas_result_distance_test00.csv};

\end{axis}
\end{tikzpicture}
\label{fig:compas_distance_x}
} \caption{Results for the COMPAS dataset: The shaded regions in (a) denotes the average DP $\pm$ std. The shaded regions in (b) denote the average WD $\pm$ std.}
\label{fig:compas}
\centering
\subfigure[Demographic parity]{
\centering
\begin{tikzpicture}
\begin{axis}[
scale=0.8,
width=0.45\linewidth,
height=100pt,
xmin=0.05,
xmax=0.5,
ymin=0,
yticklabel style={
    /pgf/number format/fixed,
    /pgf/number format/precision=2
},
xlabel={Ratio of positive decisions $\alpha$ in sampling},
ylabel={Average DP},
legend style={at={(0.5,1.1)},anchor=south,legend columns=-1, font=\footnotesize}
]

\addplot[name path=original_upper,draw=none,forget plot] table[x=ypos,y expr=\thisrow{No-Sampling}+\thisrow{Np-Sampling-std}, col sep=comma] {./csv/result_LogReg_parity.csv};
\addplot[name path=original_lower,draw=none,forget plot] table[x=ypos,y expr=\thisrow{No-Sampling}-\thisrow{Np-Sampling-std}, col sep=comma] {./csv/result_LogReg_parity.csv};
\addplot [fill=black!30,opacity=0.5,forget plot] fill between[of=original_upper and original_lower];

\addplot[name path=case_upper,draw=none,forget plot] table[x=ypos,y expr=\thisrow{Case-control}+\thisrow{Case-control-std}, col sep=comma] {./csv/result_LogReg_parity.csv};
\addplot[name path=case_lower,draw=none,forget plot] table[x=ypos,y expr=\thisrow{Case-control}-\thisrow{Case-control-std}, col sep=comma] {./csv/result_LogReg_parity.csv};
\addplot [fill=red!30,opacity=0.5,forget plot] fill between[of=case_upper and case_lower];

\addplot[name path=proposed_upper,draw=none,forget plot] table[x=ypos,y expr=\thisrow{Proposed}+\thisrow{Proposed-std}, col sep=comma] {./csv/result_LogReg_parity.csv};
\addplot[name path=proposed_lower,draw=none,forget plot] table[x=ypos,y expr=\thisrow{Proposed}-\thisrow{Proposed-std}, col sep=comma] {./csv/result_LogReg_parity.csv};
\addplot [fill=blue!30,opacity=0.5,forget plot] fill between[of=proposed_upper and proposed_lower];

\addplot[ultra thick, black, dotted] table[x=ypos,y=No-Sampling, col sep=comma]{./csv/result_LogReg_parity.csv};
\addplot[ultra thick, red, dashed] table[x=ypos,y=Case-control, col sep=comma]{./csv/result_LogReg_parity.csv};
\addplot[ultra thick, blue] table[x=ypos,y=Proposed, col sep=comma]{./csv/result_LogReg_parity.csv};

\legend{Baseline, Case-control, Stealth}

\end{axis}
\end{tikzpicture}
\label{fig:adult_parity}
}
\subfigure[Wasserstein Distance in $\Pr(x)$]{
\centering
\begin{tikzpicture}
\begin{axis}[
scale=0.8,
width=0.45\linewidth,
height=100pt,
xmin=0.05,
xmax=0.5,
xlabel={Ratio of positive decisions $\alpha$ in sampling},
ylabel={Average WD},
legend style={at={(0.37,0.97)},anchor=north}
]

\addplot[name path=original_upper,draw=none,forget plot] table[x=ypos,y expr=\thisrow{No-Sampling}+\thisrow{Np-Sampling-std}, col sep=comma] {./csv/result_LogReg_distance_test00.csv};
\addplot[name path=original_lower,draw=none,forget plot] table[x=ypos,y expr=\thisrow{No-Sampling}-\thisrow{Np-Sampling-std}, col sep=comma] {./csv/result_LogReg_distance_test00.csv};
\addplot [fill=black!30,opacity=0.5,forget plot] fill between[of=original_upper and original_lower];

\addplot[name path=case_upper,draw=none,forget plot] table[x=ypos,y expr=\thisrow{Case-control}+\thisrow{Case-control-std}, col sep=comma] {./csv/result_LogReg_distance_test00.csv};
\addplot[name path=case_lower,draw=none,forget plot] table[x=ypos,y expr=\thisrow{Case-control}-\thisrow{Case-control-std}, col sep=comma] {./csv/result_LogReg_distance_test00.csv};
\addplot [fill=red!30,opacity=0.5,forget plot] fill between[of=case_upper and case_lower];

\addplot[name path=proposed_upper,draw=none,forget plot] table[x=ypos,y expr=\thisrow{Proposed}+\thisrow{Proposed-std}, col sep=comma] {./csv/result_LogReg_distance_test00.csv};
\addplot[name path=proposed_lower,draw=none,forget plot] table[x=ypos,y expr=\thisrow{Proposed}-\thisrow{Proposed-std}, col sep=comma] {./csv/result_LogReg_distance_test00.csv};
\addplot [fill=blue!30,opacity=0.5,forget plot] fill between[of=proposed_upper and proposed_lower];

\addplot[ultra thick, black, dotted] table[x=ypos,y=No-Sampling, col sep=comma]{./csv/result_LogReg_distance_test00.csv};
\addplot[ultra thick, red, dashed] table[x=ypos,y=Case-control, col sep=comma]{./csv/result_LogReg_distance_test00.csv};
\addplot[ultra thick, blue] table[x=ypos,y=Proposed, col sep=comma]{./csv/result_LogReg_distance_test00.csv};

\end{axis}
\end{tikzpicture}
\label{fig:adult_distance_x}
} \caption{Results for the Adult dataset: The shaded regions in (a) denotes the average DP $\pm$ std. The shaded regions in (b) denote the average WD $\pm$ std.}
\label{fig:adult}
\end{figure*}

\subsection{Real-World Data: Adult}
\label{sec:adult}

As the second real-world data experiment, we used the Adult dataset~\cite{Dua:2017}.
The Adult dataset contains 48,842 records with several individual's features and their labels (high-income or low-income).
The dataset is known to include gender bias: in the dataset, while 30\% of the male have high-income, only 10\% of the female have high-income.
The DP of the dataset is therefore $0.2$.
If we naively train a classifier using the dataset, the resulting classifier inherits the bias and becomes discriminative, i.e., the classifier favors to classify males as high-income.
The goal of this experiment is to show that as if the biased classifier is fair by disclosing a part of the dataset with classifier's decision.

\medskip
\noindent
\textbf{Data \& Classifier \:}
In the data preprocessing, we converted categorical features to numerical features.\footnote{We used the implementation used in \url{https://www.kaggle.com/kost13/us-income-logistic-regression/notebook}}
We randomly split 10,000 records for the training set, 20,000 records for the test set, and the remaining 18,842 records for the referential set $D'$ for the detector.
In the experiment, we first train a classifier using the training set.
As a classifier, we used logistic regression and random forest with $100$ trees.
We labeled all the records in the test set using the trained classifier and obtained the dataset $D$ with the classifier's decision.
We then sample the subset $\bench \subseteq D$ with size $|\bench| = 2,000$ using both the stealthily biased and case-control sampling.
To reduce the DP in the sampling, we required the sampled set to satisfy $\Pr ( y=1 \mid s=1 ) \approx \Pr ( y=1 \mid s=0 ) \approx \alpha$ for a predetermined ratio of positive decisions $\alpha \in [0, 1]$.

\medskip
\noindent
\textbf{Detector \:}
We adopted the same detector as the COMPAS data experiment, who refers to the WD as the bias detection metric.

\medskip
\noindent
\textbf{Result \:}
We repeated the experiment 100 times by randomly changing the data splitting, and summarized the resultsf for logistic regression in \figurename~\ref{fig:adult}.\footnote{Here, we measured the WD on $\Pr(x)$. The WD on $\Pr(x \mid s=1)$ and $\Pr(x \mid s=0)$ can be found in Appendix~\ref{app:adult_res}. The results for random forest can be found also in Appendix~\ref{app:adult_res}.}
As the baseline, we computed the DP and the WD for randomly sampled $2,000$ sampled records from $D$, which is denoted as \emph{Baseline} in the figure.
Similar to the results of COMPAS, the figures again show the clear success of the stealthily biased sampling (R1, R2, and R3).

\section{Conclusion}

We assessed the risk of malicious decision-makers who try to deceive auditing tools, by investigating the detectability of the decision-maker's fraud.
We specifically put our focus on an auditing tool that computes a fairness metric.
To assess the (un)detectability of the fraud, we considered the biased sampling attack, where the decision-maker publishes a benchmark dataset as the evidence of his or her fairness.
In this study, we demonstrated the undetectability by explicitly constructing an algorithm, the stealthily based sampling, that can deliberately construct a fair benchmark dataset.
To derive the algorithm, we formulated the sampling problem as a Wasserstein distance minimization, which we reduced to a minimum-cost flow problem for efficient computation.
We then showed that the fraud made by the stealthily based sampling is indeed difficult to detect both theoretically and empirically.

A recent study of \cite{aivodji2019fairwashing} has shown that malicious decision-makers can rationalize their unfair decisions by generating seemingly fair explanations, which indicates that an explanation will not be effective for certifying fairnesses.
Our results indicate that passing the auditing tools will not be sufficient as the evidence of the fairness as well.
Assessing the validity of other auditing tools and mechanisms against malicious decision-makers would be essential.

Lastly, in this study, we revealed the difficulty of detecting decision-maker's fraud.
While auditing tools are getting popular, we will need additional social mechanisms that certify the reported results of these tools.
We hope that our study opens up new research directions for practical social mechanisms that can certify fairnesses.

\subsubsection*{Acknowledgments}
We would like to thank S{\'e}bastien Gambs and Ulrich A{\"i}vodji for their helpful comments.
Kazuto Fukuchi is supported by JSPS KAKENHI Grant Number JP19H04164.
Satoshi Hara is supported by JSPS KAKENHI Grant Number JP18K18106.

\bibliographystyle{plain}
\bibliography{main}

\clearpage

\appendix
\section*{Appendix}

\section{Quantitative Biasing}
\label{sec:general-biasing}

In the main body of the paper, we show a method for categorical biasing.
Here, we show that the quantitative bias can also be included.

Imagine that the sensitive attribute is quantitative (e.g., the height of a person).
In this case, the fair decision must satisfy that the expected value of the sensitive attribute in each category $y \in \mathcal{Y}$ is the same.
Let $\gamma$ be the expected value of the sensitive attribute among the dataset. 
Then, this constraint is given by
\begin{align}
   \gamma - \epsilon \le \sum_{i \in D: y_i = y} s(x_i) \mu_i \le \gamma + \epsilon, \ \ (y \in \mathcal{Y}).
\end{align}
We denote this constraint $\mu \in P(\gamma)$.
Then, the stealthily biased sampling problem is given as follows.
\begin{problem}[Stealthily biased sampling problem (Quantitative bias)]
\begin{align}
    \begin{array}{lll}
    \text{min} & W(\mu, \nu) \\
    \text{s.t.} & \mu \in P(\gamma).
    \end{array}
\end{align}
\end{problem}
By substituting the definition of the Wasserstein distance, the above problem is reduced to the following linear programming problem.
\begin{align}
\label{eq:stealthilyLP-quantitative}
    \begin{array}{lll}
    \text{min} & \sum_{i,j \in V} d(x_i, x_j) \pi_{ij} \\
    \text{s.t.} & \pi \in \Delta(\mu, \nu), \mu \in P(\gamma).
    \end{array}
\end{align}
The problem is solved using the ellipsoid method or the alternating direction method of multiplier (ADMM) with minimum-cost flow computations.

\section{Proof of Theorem \ref{thm:adv-ks-our}}
\begin{proof}
 The advantage can be rewritten as
\begin{align*}
  & \mathrm{Adv}(\Phi;\mu^K,\nu^K) \\
  &= \left|\Pr_{D \sim \mu^K}(\Phi(D) = 1) - \Pr_{D \sim \nu^K}(\Phi(D) = 1)\right| \\
  &= \left|\int \mathbb{I}(\Phi(D) = 1) \mu^K(dD) - \int \mathbb{I}(\Phi(D) = 1) \nu^K(dD)\right|.
\end{align*}
Let us approximate the function $D \to \mathbb{I}(\Phi(D) = 1)$. Let $\partial\Phi$ be the boundary across which the output of $\Phi$ is changed from $0$ or $1$ to $1$ or $0$. For $D$, let $d(D, \partial\Phi) = \inf_{D' \in \partial\Phi}d(D, D')$, where $d(D, D') = \min_{i = 1,...,K}d(x_i,x'_i)$ for $D = (x_1,...,x_K)$ and $D' = (x'_1,...,x'_K)$. For $\epsilon > 0$, define
\begin{align}
  \tilde{I}_\epsilon(D) = \begin{cases}
    \mathbb{I}(\Phi(D) = 1) & \mbox{if } d(D, \partial\Phi) \ge \epsilon, \\
    \frac{1}{2} + d(D, \partial\Phi) & \mbox{if } \Phi(D) = 1, \\
    \frac{1}{2} - d(D, \partial\Phi) & \mbox{if } \Phi(D) = 0.
  \end{cases}
\end{align}
Then, the advantage is approximated as
\begin{multline}
  \mathrm{Adv}(\Phi;\mu^K,\nu^K) \le 
  \left|\int \tilde{I}_\epsilon (d\mu^K - d\nu^K)\right| \\ + \Pr_{D \sim \mu^K}(d(D, \partial\Phi) \le \epsilon) + \Pr_{D \sim \nu^K}(d(D, \partial\Phi) \le \epsilon). \label{eq:adv-bound}
\end{multline}
By definition, $\tilde{I}_\epsilon$ is a $2\epsilon$-Lipschitz function. Hence, from Kantorovich--Rubenstein duality of 1-Wasserstein distance, we have
\begin{align}
  \left|\int \tilde{I}_\epsilon (d\mu^K - d\nu^K)\right| \le \frac{W(\mu^K,\nu^K)}{2\epsilon}.
\end{align}

Let us consider the condition under which $\Phi_{\mathrm{KS},\tau}(D) = 1$. Let $I_1,...,I_K$ be intervals such that $F_\mu(x) \in [(i-1)/K, i/K]$ for any $x \in I_i$. For an interval $I$, let $I_\epsilon = \{ x : \inf_{x' \in I}|x - x'| \le \epsilon\}$. Then, $\Phi_{\mathrm{KS},\tau}(D) = 1$ if and only if $x_{(i)} \in (I_i)_\tau$ for all $i = 1,...,K$, where $x_{(1)},...,x_{(K)}$ be the ordered samples such that $x_{(1)} \le ... \le x_{(K)}$. 

Let $x_{\mathrm{low},i} = \inf\{ x \in (I_i)_\tau \}$ and $x_{\mathrm{up},i} = \sup\{ x \in (I_i)_\tau \}$. Then, we have
\begin{align}
 & \Pr_{D \sim \nu^K}(d(D, \partial\Phi) \le \epsilon) \\
 & = \Pr_{D \sim \nu^K}\left(\forall i, \min\{|x_{(i)} - x_{\mathrm{low},i}|,|x_{(i)} - x_{\mathrm{up},i}|\} \le \epsilon \right).
\end{align}
If $\epsilon \le \tau$, the event $\forall i, \min\{|x_{(i)} - x_{\mathrm{low},i}|,|x_{(i)} - x_{\mathrm{up},i}|\} \le \epsilon$ is equivalent to either
\begin{align}
  \forall i=1,...,K, |x_{(i)} - x_{\mathrm{low},i}| \le \epsilon, \label{eq:event-low}
\end{align}
or
\begin{align}
  \forall i=1,...,K, |x_{(i)} - x_{\mathrm{up},i}| \le \epsilon, \label{eq:event-up}
\end{align}
because $x_{\mathrm{low},i} - \epsilon \ge x_{\mathrm{up},i-1} + \epsilon$ and $x_{\mathrm{up},i-1} - \epsilon \ge x_{\mathrm{low},i} + \epsilon$ for $i=2,...,K$. 

Let us derive a bound on the probability that the event Eq. \ref{eq:event-low} occurs. It can be bounded by
\begin{align}
    &\Pr(\exists i, x_1 \in B_{2\epsilon}(x_{\mathrm{low},i}))\\
    &\Pr(\exists i, x_1 \notin B_{2\epsilon}(x_{\mathrm{low},i}), x_2 \in B_{2\epsilon}(x_{\mathrm{low},i}))\\
    &\Pr(\exists i, x_1,x_2 \notin B_{2\epsilon}(x_{\mathrm{low},i}), x_3 \in B_{2\epsilon}(x_{\mathrm{low},i}))...
\end{align}
Similarly, the probability that the event Eq. \ref{eq:event-up} occurs can be bounded by
\begin{align}
    &\Pr(\exists i, x_1 \in B_{2\epsilon}(x_{\mathrm{up},i}))\\
    &\Pr(\exists i, x_1 \notin B_{2\epsilon}(x_{\mathrm{up},i}), x_2 \in B_{2\epsilon}(x_{\mathrm{up},i}))\\
    &\Pr(\exists i, x_1,x_2 \notin B_{2\epsilon}(x_{\mathrm{up},i}), x_3 \in B_{2\epsilon}(x_{\mathrm{up },i}))...
\end{align}
Hence, under $D \sim \mu^K$, we have
\begin{align}
    \Pr((\ref{eq:event-low})) \le& K!\prod_{i=1}^K\mu(B_{2\epsilon}(x_{\mathrm{low},i})), \\ 
    \Pr((\ref{eq:event-up})) \le& K!\prod_{i=1}^K\mu(B_{2\epsilon}(x_{\mathrm{up},i})). 
\end{align}
Similarly, under $D \sim \nu^K$, we have
\begin{align}
    \Pr((\ref{eq:event-low})) \le& K!\prod_{i=1}^K\nu(B_{2\epsilon}(x_{\mathrm{low},i})), \\ 
    \Pr((\ref{eq:event-up})) \le& K!\prod_{i=1}^K\nu(B_{2\epsilon}(x_{\mathrm{up},i})). 
\end{align}

Suppose $\epsilon$ is sufficiently small so that $B_{2\epsilon}(x_{\mathrm{low},i})$ for $i=1,...,K$ are distinct sets. Under Assumption \ref{asm:flatness}, it is enough if $\epsilon \le (C/K)^{1/s}/2$. Under Assumption \ref{asm:flatness}, $\nu(B_{2\epsilon}(x_{\mathrm{low},i})) \le (2\epsilon/C)^s$ and $\nu(B_{2\epsilon}(x_{\mathrm{up},i})) \le (2\epsilon/C)^s$ for any $i=1,...,K$. Note that the probability mass that is moved from $\nu$ to $\mu$ is at most $\mathrm{TV}(\kappa,\kappa')$. There exist $\alpha_i \ge 0$ satisfying $\sum_{i=1}^K\alpha_i = 1$ such that for any $i = 1,...,K$,
\begin{align}
 \mu(B_{2\epsilon}(x_{\mathrm{low},i})) \le \nu(B_{2\epsilon}(x_{\mathrm{low},i})) + \alpha_i\mathrm{TV}(\kappa,\kappa'). \nonumber
\end{align}
Thus, we have
\begin{align}
   \Pr((\ref{eq:event-low})) \le& K!\prod_{i=1}^K\left(\nu(B_{2\epsilon}(x_{\mathrm{low},i})) + \alpha_i\mathrm{TV}(\kappa,\kappa')\right) \nonumber\\
   \le& K!\left(\left(\frac{2\epsilon}{C}\right)^s + \frac{\mathrm{TV}(\kappa,\kappa')}{K}\right)^K.
\end{align}
The same bound holds for $\Pr((\ref{eq:event-up}))$. Substituting these bounds into Eq. \ref{eq:adv-bound} yields 
\begin{align}
  \mathrm{Adv}(\Phi;\mu^K,\nu^K) \le & \frac{W(\mu^K,\nu^K)}{2\epsilon} \nonumber \\
    & + 4K!\left(\left(\frac{2\epsilon}{C}\right)^s + \frac{\mathrm{TV}(\kappa,\kappa')}{K}\right)^K.
\end{align}
Setting $\epsilon = (C/K)^{1/s}/2$ yields the claim.
\end{proof}

\section{Bootstrap-type Estimator}
\label{app:bootstrap}

In real-world data, the computational scalability of the stealthily biased sampling can be a bottleneck: it requires $\tilde{O}(N^{2.5})$ time for the dataset of size $N$ using a general minimum-cost flow solver.
Empirically, we observed that solving the problem for $N \ge 10,000$ tends to be computationally prohibitive.
Here, we consider a bootstrap-type estimator to bypass the prohibitive computation.
In the every round of the bootstrap, we sample $N'$ points $D'$ out of the $N$ points in the dataset $D$, i.e., $D' \subseteq D$ and $|D'| = N'$.
We solve the minimum-cost flow problem only on the subset $D'$ and obtain a measure $\mu'$.
Finally, we average the measures $\mu'$ obtained in the every bootstrap round as the estimated measure $\mu$.
If we take $N'$ in a reasonable size (e.g. around a few thousands), this estimation is sufficiently efficient, without much loss on the estimation accuracy.
The bootstrap step can be easily parallelized to speed up the computation further.

In the Adult data experiment in Section~\ref{sec:experiments}, we set the sample size $N'=4,000$ and the number of bootstrap steps to be 30.

\section{Synthetic Example: Additional Results}
\label{app:synthetic}

Here, we present additional results for the synthetic data experiment in Section~\ref{sec:synthetic}.

\subsection{Results in higher dimensional settings}
\label{app:synthetic_deterministic}

We conducted additional experiments by changing the dimensionality $d$ of the feature $x$ to $d=2, 5$, and $10$.
The results are shown in Figures~\ref{fig:app_synthetic_deterministic_d02}, \ref{fig:app_synthetic_deterministic_d05}, and \ref{fig:app_synthetic_deterministic_d10}.
Those results also support our key observations R1, R2, and R3 in Section~\ref{sec:experiments}.

\begin{figure*}[t]
\centering
\subfigure[Demographic parity]{
\centering
\begin{tikzpicture}
\begin{axis}[
scale=0.8,
width=0.45\linewidth,
height=120pt,
xmin=0.4,
xmax=0.8,
ymax=0.15,
yticklabel style={
    /pgf/number format/fixed,
    /pgf/number format/precision=2
},
xlabel={Ratio of positive decisions $\alpha$ in sampling},
ylabel={Average DP},
legend style={at={(0.5,1.1)},anchor=south,legend columns=-1, font=\footnotesize}
]

\addplot[name path=case_upper,draw=none,forget plot] table[x=a,y expr=\thisrow{Case-control}+\thisrow{Case-control-std}, col sep=comma] {./csv/parity_deterministic_d002_q20.csv};
\addplot[name path=case_lower,draw=none,forget plot] table[x=a,y expr=\thisrow{Case-control}-\thisrow{Case-control-std}, col sep=comma] {./csv/parity_deterministic_d002_q20.csv};
\addplot [fill=red!30,opacity=0.5,forget plot] fill between[of=case_upper and case_lower];

\addplot[name path=proposed_upper,draw=none,forget plot] table[x=a,y expr=\thisrow{Proposed}+\thisrow{Proposed-std}, col sep=comma] {./csv/parity_deterministic_d002_q20.csv};
\addplot[name path=proposed_lower,draw=none,forget plot] table[x=a,y expr=\thisrow{Proposed}-\thisrow{Proposed-std}, col sep=comma] {./csv/parity_deterministic_d002_q20.csv};
\addplot [fill=blue!30,opacity=0.5,forget plot] fill between[of=proposed_upper and proposed_lower];

\addplot[ultra thick, red, dashed] table[x=a,y=Case-control, col sep=comma]{./csv/parity_deterministic_d002_q20.csv};
\addplot[ultra thick, blue] table[x=a,y=Proposed, col sep=comma]{./csv/parity_deterministic_d002_q20.csv};

\legend{Case-control, Stealth}

\end{axis}
\end{tikzpicture}
\label{fig:synthetic_deterministic_parity_d02}
}
\subfigure[Ratio of rejection in test $\Pr(x)$]{
\centering
\begin{tikzpicture}
\begin{axis}[
scale=0.8,
width=0.45\linewidth,
height=120pt,
xmin=0.4,
xmax=0.8,
xlabel={Ratio of positive decisions $\alpha$ in sampling},
ylabel={Ratio of rejection},
legend style={at={(0.47,0.97)},anchor=north}
]

\addplot[name path=case_upper,draw=none,forget plot] table[x=a,y expr=\thisrow{Case-control}+\thisrow{Case-control-conf}, col sep=comma] {./csv/reject_deterministic_d002_q20_test0.csv};
\addplot[name path=case_lower,draw=none,forget plot] table[x=a,y expr=\thisrow{Case-control}-\thisrow{Case-control-conf}, col sep=comma] {./csv/reject_deterministic_d002_q20_test0.csv};
\addplot [fill=red!30,opacity=0.5,forget plot] fill between[of=case_upper and case_lower];

\addplot[name path=proposed_upper,draw=none,forget plot] table[x=a,y expr=\thisrow{Proposed}+\thisrow{Proposed-conf}, col sep=comma] {./csv/reject_deterministic_d002_q20_test0.csv};
\addplot[name path=proposed_lower,draw=none,forget plot] table[x=a,y expr=\thisrow{Proposed}-\thisrow{Proposed-conf}, col sep=comma] {./csv/reject_deterministic_d002_q20_test0.csv};
\addplot [fill=blue!30,opacity=0.5,forget plot] fill between[of=proposed_upper and proposed_lower];

\addplot[ultra thick, red, dashed] table[x=a,y=Case-control, col sep=comma]{./csv/reject_deterministic_d002_q20_test0.csv};
\addplot[ultra thick, blue] table[x=a,y=Proposed, col sep=comma]{./csv/reject_deterministic_d002_q20_test0.csv};
\addplot[ultra thick, black, dotted, domain=0:1] {0.05};

\end{axis}
\end{tikzpicture}
\label{fig:synthetic_deterministic_reject_x_d02}
}
\subfigure[Ratio of rejection in test $\Pr(x|s=1)$]{
\centering
\begin{tikzpicture}
\begin{axis}[
scale=0.8,
width=0.45\linewidth,
height=120pt,
xmin=0.4,
xmax=0.8,
xlabel={Ratio of positive decisions $\alpha$ in sampling},
ylabel={Ratio of rejection},
legend pos=north east
]

\addplot[name path=case_upper,draw=none,forget plot] table[x=a,y expr=\thisrow{Case-control}+\thisrow{Case-control-conf}, col sep=comma] {./csv/reject_deterministic_d002_q20_test1.csv};
\addplot[name path=case_lower,draw=none,forget plot] table[x=a,y expr=\thisrow{Case-control}-\thisrow{Case-control-conf}, col sep=comma] {./csv/reject_deterministic_d002_q20_test1.csv};
\addplot [fill=red!30,opacity=0.5,forget plot] fill between[of=case_upper and case_lower];

\addplot[name path=proposed_upper,draw=none,forget plot] table[x=a,y expr=\thisrow{Proposed}+\thisrow{Proposed-conf}, col sep=comma] {./csv/reject_deterministic_d002_q20_test1.csv};
\addplot[name path=proposed_lower,draw=none,forget plot] table[x=a,y expr=\thisrow{Proposed}-\thisrow{Proposed-conf}, col sep=comma] {./csv/reject_deterministic_d002_q20_test1.csv};
\addplot [fill=blue!30,opacity=0.5,forget plot] fill between[of=proposed_upper and proposed_lower];

\addplot[ultra thick, red, dashed] table[x=a,y=Case-control, col sep=comma]{./csv/reject_deterministic_d002_q20_test1.csv};
\addplot[ultra thick, blue] table[x=a,y=Proposed, col sep=comma]{./csv/reject_deterministic_d002_q20_test1.csv};
\addplot[ultra thick, black, dotted, domain=0:1] {0.05};

\end{axis}
\end{tikzpicture}
\label{fig:synthetic_deterministic_reject_x_s1_d02}
}
\subfigure[Ratio of rejection in test $\Pr(x|s=0)$]{
\centering
\begin{tikzpicture}
\begin{axis}[
scale=0.8,
width=0.45\linewidth,
height=120pt,
xmin=0.4,
xmax=0.8,
xlabel={Ratio of positive decisions $\alpha$ in sampling},
ylabel={Ratio of rejection},
legend pos=north west
]

\addplot[name path=case_upper,draw=none,forget plot] table[x=a,y expr=\thisrow{Case-control}+\thisrow{Case-control-conf}, col sep=comma] {./csv/reject_deterministic_d002_q20_test2.csv};
\addplot[name path=case_lower,draw=none,forget plot] table[x=a,y expr=\thisrow{Case-control}-\thisrow{Case-control-conf}, col sep=comma] {./csv/reject_deterministic_d002_q20_test2.csv};
\addplot [fill=red!30,opacity=0.5,forget plot] fill between[of=case_upper and case_lower];

\addplot[name path=proposed_upper,draw=none,forget plot] table[x=a,y expr=\thisrow{Proposed}+\thisrow{Proposed-conf}, col sep=comma] {./csv/reject_deterministic_d002_q20_test2.csv};
\addplot[name path=proposed_lower,draw=none,forget plot] table[x=a,y expr=\thisrow{Proposed}-\thisrow{Proposed-conf}, col sep=comma] {./csv/reject_deterministic_d002_q20_test2.csv};
\addplot [fill=blue!30,opacity=0.5,forget plot] fill between[of=proposed_upper and proposed_lower];

\addplot[ultra thick, red, dashed] table[x=a,y=Case-control, col sep=comma]{./csv/reject_deterministic_d002_q20_test2.csv};
\addplot[ultra thick, blue] table[x=a,y=Proposed, col sep=comma]{./csv/reject_deterministic_d002_q20_test2.csv};
\addplot[ultra thick, black, dotted, domain=0:1] {0.05};

\end{axis}
\end{tikzpicture}
\label{fig:synthetic_deterministic_reject_x_s0_d02}
} \caption{[$d=2$] The results for the decision-maker with $a=0.2$: The shaded regions in (a) denotes the average DP $\pm$ std. The shaded regions in (b), (c), and (d) denote 95\% confidence intervals. The dotted line in (b), (c), and (d) denotes the significance level $0.05$.}
\label{fig:app_synthetic_deterministic_d02}
\vspace{24pt}
\centering
\subfigure[Demographic parity]{
\centering
\begin{tikzpicture}
\begin{axis}[
scale=0.8,
width=0.45\linewidth,
height=120pt,
xmin=0.4,
xmax=0.8,
ymax=0.15,
yticklabel style={
    /pgf/number format/fixed,
    /pgf/number format/precision=2
},
xlabel={Ratio of positive decisions $\alpha$ in sampling},
ylabel={Average DP},
legend style={at={(0.5,1.1)},anchor=south,legend columns=-1, font=\footnotesize}
]

\addplot[name path=case_upper,draw=none,forget plot] table[x=a,y expr=\thisrow{Case-control}+\thisrow{Case-control-std}, col sep=comma] {./csv/parity_deterministic_d005_q20.csv};
\addplot[name path=case_lower,draw=none,forget plot] table[x=a,y expr=\thisrow{Case-control}-\thisrow{Case-control-std}, col sep=comma] {./csv/parity_deterministic_d005_q20.csv};
\addplot [fill=red!30,opacity=0.5,forget plot] fill between[of=case_upper and case_lower];

\addplot[name path=proposed_upper,draw=none,forget plot] table[x=a,y expr=\thisrow{Proposed}+\thisrow{Proposed-std}, col sep=comma] {./csv/parity_deterministic_d005_q20.csv};
\addplot[name path=proposed_lower,draw=none,forget plot] table[x=a,y expr=\thisrow{Proposed}-\thisrow{Proposed-std}, col sep=comma] {./csv/parity_deterministic_d005_q20.csv};
\addplot [fill=blue!30,opacity=0.5,forget plot] fill between[of=proposed_upper and proposed_lower];

\addplot[ultra thick, red, dashed] table[x=a,y=Case-control, col sep=comma]{./csv/parity_deterministic_d005_q20.csv};
\addplot[ultra thick, blue] table[x=a,y=Proposed, col sep=comma]{./csv/parity_deterministic_d005_q20.csv};

\legend{Case-control, Stealth}

\end{axis}
\end{tikzpicture}
\label{fig:synthetic_deterministic_parity_d05}
}
\subfigure[Ratio of rejection in test $\Pr(x)$]{
\centering
\begin{tikzpicture}
\begin{axis}[
scale=0.8,
width=0.45\linewidth,
height=120pt,
xmin=0.4,
xmax=0.8,
xlabel={Ratio of positive decisions $\alpha$ in sampling},
ylabel={Ratio of rejection},
legend style={at={(0.47,0.97)},anchor=north}
]

\addplot[name path=case_upper,draw=none,forget plot] table[x=a,y expr=\thisrow{Case-control}+\thisrow{Case-control-conf}, col sep=comma] {./csv/reject_deterministic_d005_q20_test0.csv};
\addplot[name path=case_lower,draw=none,forget plot] table[x=a,y expr=\thisrow{Case-control}-\thisrow{Case-control-conf}, col sep=comma] {./csv/reject_deterministic_d005_q20_test0.csv};
\addplot [fill=red!30,opacity=0.5,forget plot] fill between[of=case_upper and case_lower];

\addplot[name path=proposed_upper,draw=none,forget plot] table[x=a,y expr=\thisrow{Proposed}+\thisrow{Proposed-conf}, col sep=comma] {./csv/reject_deterministic_d005_q20_test0.csv};
\addplot[name path=proposed_lower,draw=none,forget plot] table[x=a,y expr=\thisrow{Proposed}-\thisrow{Proposed-conf}, col sep=comma] {./csv/reject_deterministic_d005_q20_test0.csv};
\addplot [fill=blue!30,opacity=0.5,forget plot] fill between[of=proposed_upper and proposed_lower];

\addplot[ultra thick, red, dashed] table[x=a,y=Case-control, col sep=comma]{./csv/reject_deterministic_d005_q20_test0.csv};
\addplot[ultra thick, blue] table[x=a,y=Proposed, col sep=comma]{./csv/reject_deterministic_d005_q20_test0.csv};
\addplot[ultra thick, black, dotted, domain=0:1] {0.05};

\end{axis}
\end{tikzpicture}
\label{fig:synthetic_deterministic_reject_x_d05}
}
\subfigure[Ratio of rejection in test $\Pr(x|s=1)$]{
\centering
\begin{tikzpicture}
\begin{axis}[
scale=0.8,
width=0.45\linewidth,
height=120pt,
xmin=0.4,
xmax=0.8,
xlabel={Ratio of positive decisions $\alpha$ in sampling},
ylabel={Ratio of rejection},
legend pos=north east
]

\addplot[name path=case_upper,draw=none,forget plot] table[x=a,y expr=\thisrow{Case-control}+\thisrow{Case-control-conf}, col sep=comma] {./csv/reject_deterministic_d005_q20_test1.csv};
\addplot[name path=case_lower,draw=none,forget plot] table[x=a,y expr=\thisrow{Case-control}-\thisrow{Case-control-conf}, col sep=comma] {./csv/reject_deterministic_d005_q20_test1.csv};
\addplot [fill=red!30,opacity=0.5,forget plot] fill between[of=case_upper and case_lower];

\addplot[name path=proposed_upper,draw=none,forget plot] table[x=a,y expr=\thisrow{Proposed}+\thisrow{Proposed-conf}, col sep=comma] {./csv/reject_deterministic_d005_q20_test1.csv};
\addplot[name path=proposed_lower,draw=none,forget plot] table[x=a,y expr=\thisrow{Proposed}-\thisrow{Proposed-conf}, col sep=comma] {./csv/reject_deterministic_d005_q20_test1.csv};
\addplot [fill=blue!30,opacity=0.5,forget plot] fill between[of=proposed_upper and proposed_lower];

\addplot[ultra thick, red, dashed] table[x=a,y=Case-control, col sep=comma]{./csv/reject_deterministic_d005_q20_test1.csv};
\addplot[ultra thick, blue] table[x=a,y=Proposed, col sep=comma]{./csv/reject_deterministic_d005_q20_test1.csv};
\addplot[ultra thick, black, dotted, domain=0:1] {0.05};

\end{axis}
\end{tikzpicture}
\label{fig:synthetic_deterministic_reject_x_s1_d05}
}
\subfigure[Ratio of rejection in test $\Pr(x|s=0)$]{
\centering
\begin{tikzpicture}
\begin{axis}[
scale=0.8,
width=0.45\linewidth,
height=120pt,
xmin=0.4,
xmax=0.8,
xlabel={Ratio of positive decisions $\alpha$ in sampling},
ylabel={Ratio of rejection},
legend pos=north west
]

\addplot[name path=case_upper,draw=none,forget plot] table[x=a,y expr=\thisrow{Case-control}+\thisrow{Case-control-conf}, col sep=comma] {./csv/reject_deterministic_d005_q20_test2.csv};
\addplot[name path=case_lower,draw=none,forget plot] table[x=a,y expr=\thisrow{Case-control}-\thisrow{Case-control-conf}, col sep=comma] {./csv/reject_deterministic_d005_q20_test2.csv};
\addplot [fill=red!30,opacity=0.5,forget plot] fill between[of=case_upper and case_lower];

\addplot[name path=proposed_upper,draw=none,forget plot] table[x=a,y expr=\thisrow{Proposed}+\thisrow{Proposed-conf}, col sep=comma] {./csv/reject_deterministic_d005_q20_test2.csv};
\addplot[name path=proposed_lower,draw=none,forget plot] table[x=a,y expr=\thisrow{Proposed}-\thisrow{Proposed-conf}, col sep=comma] {./csv/reject_deterministic_d005_q20_test2.csv};
\addplot [fill=blue!30,opacity=0.5,forget plot] fill between[of=proposed_upper and proposed_lower];

\addplot[ultra thick, red, dashed] table[x=a,y=Case-control, col sep=comma]{./csv/reject_deterministic_d005_q20_test2.csv};
\addplot[ultra thick, blue] table[x=a,y=Proposed, col sep=comma]{./csv/reject_deterministic_d005_q20_test2.csv};
\addplot[ultra thick, black, dotted, domain=0:1] {0.05};

\end{axis}
\end{tikzpicture}
\label{fig:synthetic_deterministic_reject_x_s0_d05}
} \caption{[$d=5$] The results for the decision-maker with $a=0.2$: The shaded regions in (a) denotes the average DP $\pm$ std. The shaded regions in (b), (c), and (d) denote 95\% confidence intervals. The dotted line in (b), (c), and (d) denotes the significance level $0.05$.}
\label{fig:app_synthetic_deterministic_d05}
\end{figure*}

\begin{figure*}[t]
\centering
\subfigure[Demographic parity]{
\centering
\begin{tikzpicture}
\begin{axis}[
scale=0.8,
width=0.45\linewidth,
height=120pt,
xmin=0.4,
xmax=0.8,
ymax=0.15,
yticklabel style={
    /pgf/number format/fixed,
    /pgf/number format/precision=2
},
xlabel={Ratio of positive decisions $\alpha$ in sampling},
ylabel={Average DP},
legend style={at={(0.5,1.1)},anchor=south,legend columns=-1, font=\footnotesize}
]

\addplot[name path=case_upper,draw=none,forget plot] table[x=a,y expr=\thisrow{Case-control}+\thisrow{Case-control-std}, col sep=comma] {./csv/parity_deterministic_d010_q20.csv};
\addplot[name path=case_lower,draw=none,forget plot] table[x=a,y expr=\thisrow{Case-control}-\thisrow{Case-control-std}, col sep=comma] {./csv/parity_deterministic_d010_q20.csv};
\addplot [fill=red!30,opacity=0.5,forget plot] fill between[of=case_upper and case_lower];

\addplot[name path=proposed_upper,draw=none,forget plot] table[x=a,y expr=\thisrow{Proposed}+\thisrow{Proposed-std}, col sep=comma] {./csv/parity_deterministic_d010_q20.csv};
\addplot[name path=proposed_lower,draw=none,forget plot] table[x=a,y expr=\thisrow{Proposed}-\thisrow{Proposed-std}, col sep=comma] {./csv/parity_deterministic_d010_q20.csv};
\addplot [fill=blue!30,opacity=0.5,forget plot] fill between[of=proposed_upper and proposed_lower];

\addplot[ultra thick, red, dashed] table[x=a,y=Case-control, col sep=comma]{./csv/parity_deterministic_d010_q20.csv};
\addplot[ultra thick, blue] table[x=a,y=Proposed, col sep=comma]{./csv/parity_deterministic_d010_q20.csv};

\legend{Case-control, Stealth}

\end{axis}
\end{tikzpicture}
\label{fig:synthetic_deterministic_parity_d10}
}
\subfigure[Ratio of rejection in test $\Pr(x)$]{
\centering
\begin{tikzpicture}
\begin{axis}[
scale=0.8,
width=0.45\linewidth,
height=120pt,
xmin=0.4,
xmax=0.8,
xlabel={Ratio of positive decisions $\alpha$ in sampling},
ylabel={Ratio of rejection},
legend style={at={(0.47,0.97)},anchor=north}
]

\addplot[name path=case_upper,draw=none,forget plot] table[x=a,y expr=\thisrow{Case-control}+\thisrow{Case-control-conf}, col sep=comma] {./csv/reject_deterministic_d010_q20_test0.csv};
\addplot[name path=case_lower,draw=none,forget plot] table[x=a,y expr=\thisrow{Case-control}-\thisrow{Case-control-conf}, col sep=comma] {./csv/reject_deterministic_d010_q20_test0.csv};
\addplot [fill=red!30,opacity=0.5,forget plot] fill between[of=case_upper and case_lower];

\addplot[name path=proposed_upper,draw=none,forget plot] table[x=a,y expr=\thisrow{Proposed}+\thisrow{Proposed-conf}, col sep=comma] {./csv/reject_deterministic_d010_q20_test0.csv};
\addplot[name path=proposed_lower,draw=none,forget plot] table[x=a,y expr=\thisrow{Proposed}-\thisrow{Proposed-conf}, col sep=comma] {./csv/reject_deterministic_d010_q20_test0.csv};
\addplot [fill=blue!30,opacity=0.5,forget plot] fill between[of=proposed_upper and proposed_lower];

\addplot[ultra thick, red, dashed] table[x=a,y=Case-control, col sep=comma]{./csv/reject_deterministic_d010_q20_test0.csv};
\addplot[ultra thick, blue] table[x=a,y=Proposed, col sep=comma]{./csv/reject_deterministic_d010_q20_test0.csv};
\addplot[ultra thick, black, dotted, domain=0:1] {0.05};

\end{axis}
\end{tikzpicture}
\label{fig:synthetic_deterministic_reject_x_d10}
}
\subfigure[Ratio of rejection in test $\Pr(x|s=1)$]{
\centering
\begin{tikzpicture}
\begin{axis}[
scale=0.8,
width=0.45\linewidth,
height=120pt,
xmin=0.4,
xmax=0.8,
xlabel={Ratio of positive decisions $\alpha$ in sampling},
ylabel={Ratio of rejection},
legend pos=north east
]

\addplot[name path=case_upper,draw=none,forget plot] table[x=a,y expr=\thisrow{Case-control}+\thisrow{Case-control-conf}, col sep=comma] {./csv/reject_deterministic_d010_q20_test1.csv};
\addplot[name path=case_lower,draw=none,forget plot] table[x=a,y expr=\thisrow{Case-control}-\thisrow{Case-control-conf}, col sep=comma] {./csv/reject_deterministic_d010_q20_test1.csv};
\addplot [fill=red!30,opacity=0.5,forget plot] fill between[of=case_upper and case_lower];

\addplot[name path=proposed_upper,draw=none,forget plot] table[x=a,y expr=\thisrow{Proposed}+\thisrow{Proposed-conf}, col sep=comma] {./csv/reject_deterministic_d010_q20_test1.csv};
\addplot[name path=proposed_lower,draw=none,forget plot] table[x=a,y expr=\thisrow{Proposed}-\thisrow{Proposed-conf}, col sep=comma] {./csv/reject_deterministic_d010_q20_test1.csv};
\addplot [fill=blue!30,opacity=0.5,forget plot] fill between[of=proposed_upper and proposed_lower];

\addplot[ultra thick, red, dashed] table[x=a,y=Case-control, col sep=comma]{./csv/reject_deterministic_d010_q20_test1.csv};
\addplot[ultra thick, blue] table[x=a,y=Proposed, col sep=comma]{./csv/reject_deterministic_d010_q20_test1.csv};
\addplot[ultra thick, black, dotted, domain=0:1] {0.05};

\end{axis}
\end{tikzpicture}
\label{fig:synthetic_deterministic_reject_x_s1_d10}
}
\subfigure[Ratio of rejection in test $\Pr(x|s=0)$]{
\centering
\begin{tikzpicture}
\begin{axis}[
scale=0.8,
width=0.45\linewidth,
height=120pt,
xmin=0.4,
xmax=0.8,
xlabel={Ratio of positive decisions $\alpha$ in sampling},
ylabel={Ratio of rejection},
legend pos=north west
]

\addplot[name path=case_upper,draw=none,forget plot] table[x=a,y expr=\thisrow{Case-control}+\thisrow{Case-control-conf}, col sep=comma] {./csv/reject_deterministic_d010_q20_test2.csv};
\addplot[name path=case_lower,draw=none,forget plot] table[x=a,y expr=\thisrow{Case-control}-\thisrow{Case-control-conf}, col sep=comma] {./csv/reject_deterministic_d010_q20_test2.csv};
\addplot [fill=red!30,opacity=0.5,forget plot] fill between[of=case_upper and case_lower];

\addplot[name path=proposed_upper,draw=none,forget plot] table[x=a,y expr=\thisrow{Proposed}+\thisrow{Proposed-conf}, col sep=comma] {./csv/reject_deterministic_d010_q20_test2.csv};
\addplot[name path=proposed_lower,draw=none,forget plot] table[x=a,y expr=\thisrow{Proposed}-\thisrow{Proposed-conf}, col sep=comma] {./csv/reject_deterministic_d010_q20_test2.csv};
\addplot [fill=blue!30,opacity=0.5,forget plot] fill between[of=proposed_upper and proposed_lower];

\addplot[ultra thick, red, dashed] table[x=a,y=Case-control, col sep=comma]{./csv/reject_deterministic_d010_q20_test2.csv};
\addplot[ultra thick, blue] table[x=a,y=Proposed, col sep=comma]{./csv/reject_deterministic_d010_q20_test2.csv};
\addplot[ultra thick, black, dotted, domain=0:1] {0.05};

\end{axis}
\end{tikzpicture}
\label{fig:synthetic_deterministic_reject_x_s0_d10}
} \caption{[$d=10$] The results for the decision-maker with $a=0.2$: The shaded regions in (a) denotes the average DP $\pm$ std. The shaded regions in (b), (c), and (d) denote 95\% confidence intervals. The dotted line in (b), (c), and (d) denotes the significance level $0.05$.}
\label{fig:app_synthetic_deterministic_d10}
\end{figure*}

\subsection{Stochastic decision-maker}
\label{app:synthetic_stochastic}

Here, we demonstrate that the stochastic decision-maker can be far evil than the deterministic decision-maker.
That is, by using the stealthily biased sampling, the stochastic decision-maker can choose the ratio of positive decisions $\alpha$ in sampling almost arbitrary.
We show that even if the decision-maker make such an intense sampling, the detector cannot detect the fraud of the decision-maker.
For the experiment, instead of the criteria~\eqref{eq:simple_decision}, we consider the decision-maker that makes the decision based on the probability
\begin{align}
    \Pr ( y=1 \mid x, s ) = x_1 + b s ,
    \label{eq:stochastic_decision}
\end{align}
where $b \ge 0$ is a constant.
The setup of the experiment is the same as Section~\ref{sec:synthetic} except for the criteria of the decision-maker.

\medskip
\noindent
\textbf{Result \:}
We set the parameters in \eqref{eq:stochastic_decision} with $b=0.2$, which makes the DP of the decision-maker to be $0.2$.
We run the experiment 100 times while requiring $\Pr ( y=1 \mid s=1 ) \approx \Pr ( y=1 \mid s=0 ) \approx \alpha$ for a predetermined ratio of positive decisions $\alpha \in [0, 1]$.
The results are summarized in \figurename~\ref{fig:app_synthetic_stochastic}.
The figures indicate similar tendencies as we have already observed in the deterministic decision-maker in Section~\ref{sec:synthetic}.
One significant difference is that the stealthily biased sampling marked low rejection rates in a very wide range of $\alpha$ in sampling.
In the deterministic case, as shown in \figurename~\ref{fig:synthetic_deterministic}, $\alpha=0.6$ was the only choice for the decision-maker to pretend to be fair safely.
In the stochastic case, the decision-maker can choose arbitrary $\alpha$ between $0.25$ and $0.95$.
The results indicate that, even for such an intense sampling, the detector cannot detect the fraud of the decision-maker.

\begin{figure*}[t]
\centering
\subfigure[Demographic parity]{
\centering
\begin{tikzpicture}
\begin{axis}[
scale=0.8,
width=0.45\linewidth,
height=120pt,
xmin=0.05,
xmax=0.95,
ymax=0.15,
yticklabel style={
    /pgf/number format/fixed,
    /pgf/number format/precision=2
},
xlabel={Ratio of positive decisions $\alpha$ in sampling},
ylabel={Average DP},
legend style={at={(0.5,1.1)},anchor=south,legend columns=-1, font=\footnotesize}
]

\addplot[name path=case_upper,draw=none,forget plot] table[x=a,y expr=\thisrow{Case-control}+\thisrow{Case-control-std}, col sep=comma] {./csv/parity_stochastic_d001_q20.csv};
\addplot[name path=case_lower,draw=none,forget plot] table[x=a,y expr=\thisrow{Case-control}-\thisrow{Case-control-std}, col sep=comma] {./csv/parity_stochastic_d001_q20.csv};
\addplot [fill=red!30,opacity=0.5,forget plot] fill between[of=case_upper and case_lower];

\addplot[name path=proposed_upper,draw=none,forget plot] table[x=a,y expr=\thisrow{Proposed}+\thisrow{Proposed-std}, col sep=comma] {./csv/parity_stochastic_d001_q20.csv};
\addplot[name path=proposed_lower,draw=none,forget plot] table[x=a,y expr=\thisrow{Proposed}-\thisrow{Proposed-std}, col sep=comma] {./csv/parity_stochastic_d001_q20.csv};
\addplot [fill=blue!30,opacity=0.5,forget plot] fill between[of=proposed_upper and proposed_lower];

\addplot[ultra thick, red, dashed] table[x=a,y=Case-control, col sep=comma]{./csv/parity_stochastic_d001_q20.csv};
\addplot[ultra thick, blue] table[x=a,y=Proposed, col sep=comma]{./csv/parity_stochastic_d001_q20.csv};

\legend{Case-control, Stealth}

\end{axis}
\end{tikzpicture}
\label{fig:synthetic_stochastic_parity}
}
\subfigure[Ratio of rejection in test $\Pr(x)$]{
\centering
\begin{tikzpicture}
\begin{axis}[
scale=0.8,
width=0.45\linewidth,
height=120pt,
xmin=0.05,
xmax=0.95,
xlabel={Ratio of positive decisions $\alpha$ in sampling},
ylabel={Ratio of rejection},
legend style={at={(0.67,0.97)},anchor=north}
]

\addplot[name path=case_upper,draw=none,forget plot] table[x=a,y expr=\thisrow{Case-control}+\thisrow{Case-control-conf}, col sep=comma] {./csv/reject_stochastic_d001_q20_test0.csv};
\addplot[name path=case_lower,draw=none,forget plot] table[x=a,y expr=\thisrow{Case-control}-\thisrow{Case-control-conf}, col sep=comma] {./csv/reject_stochastic_d001_q20_test0.csv};
\addplot [fill=red!30,opacity=0.5,forget plot] fill between[of=case_upper and case_lower];

\addplot[name path=proposed_upper,draw=none,forget plot] table[x=a,y expr=\thisrow{Proposed}+\thisrow{Proposed-conf}, col sep=comma] {./csv/reject_stochastic_d001_q20_test0.csv};
\addplot[name path=proposed_lower,draw=none,forget plot] table[x=a,y expr=\thisrow{Proposed}-\thisrow{Proposed-conf}, col sep=comma] {./csv/reject_stochastic_d001_q20_test0.csv};
\addplot [fill=blue!30,opacity=0.5,forget plot] fill between[of=proposed_upper and proposed_lower];

\addplot[ultra thick, red, dashed] table[x=a,y=Case-control, col sep=comma]{./csv/reject_stochastic_d001_q20_test0.csv};
\addplot[ultra thick, blue] table[x=a,y=Proposed, col sep=comma]{./csv/reject_stochastic_d001_q20_test0.csv};
\addplot[ultra thick, black, dotted, domain=0:1] {0.05};

\end{axis}
\end{tikzpicture}
\label{fig:synthetic_stochastic_reject_x}
}
\subfigure[Ratio of rejection in test $\Pr(x|s=1)$]{
\centering
\begin{tikzpicture}
\begin{axis}[
scale=0.8,
width=0.45\linewidth,
height=120pt,
xmin=0.05,
xmax=0.95,
xlabel={Ratio of positive decisions $\alpha$ in sampling},
ylabel={Ratio of rejection},
legend pos=north east
]

\addplot[name path=case_upper,draw=none,forget plot] table[x=a,y expr=\thisrow{Case-control}+\thisrow{Case-control-conf}, col sep=comma] {./csv/reject_stochastic_d001_q20_test1.csv};
\addplot[name path=case_lower,draw=none,forget plot] table[x=a,y expr=\thisrow{Case-control}-\thisrow{Case-control-conf}, col sep=comma] {./csv/reject_stochastic_d001_q20_test1.csv};
\addplot [fill=red!30,opacity=0.5,forget plot] fill between[of=case_upper and case_lower];

\addplot[name path=proposed_upper,draw=none,forget plot] table[x=a,y expr=\thisrow{Proposed}+\thisrow{Proposed-conf}, col sep=comma] {./csv/reject_stochastic_d001_q20_test1.csv};
\addplot[name path=proposed_lower,draw=none,forget plot] table[x=a,y expr=\thisrow{Proposed}-\thisrow{Proposed-conf}, col sep=comma] {./csv/reject_stochastic_d001_q20_test1.csv};
\addplot [fill=blue!30,opacity=0.5,forget plot] fill between[of=proposed_upper and proposed_lower];

\addplot[ultra thick, red, dashed] table[x=a,y=Case-control, col sep=comma]{./csv/reject_stochastic_d001_q20_test1.csv};
\addplot[ultra thick, blue] table[x=a,y=Proposed, col sep=comma]{./csv/reject_stochastic_d001_q20_test1.csv};
\addplot[ultra thick, black, dotted, domain=0:1] {0.05};

\end{axis}
\end{tikzpicture}
\label{fig:synthetic_stochastic_reject_x_s1}
}
\subfigure[Ratio of rejection in test $\Pr(x|s=0)$]{
\centering
\begin{tikzpicture}
\begin{axis}[
scale=0.8,
width=0.45\linewidth,
height=120pt,
xmin=0.05,
xmax=0.95,
xlabel={Ratio of positive decisions $\alpha$ in sampling},
ylabel={Ratio of rejection},
legend pos=north west
]

\addplot[name path=case_upper,draw=none,forget plot] table[x=a,y expr=\thisrow{Case-control}+\thisrow{Case-control-conf}, col sep=comma] {./csv/reject_stochastic_d001_q20_test2.csv};
\addplot[name path=case_lower,draw=none,forget plot] table[x=a,y expr=\thisrow{Case-control}-\thisrow{Case-control-conf}, col sep=comma] {./csv/reject_stochastic_d001_q20_test2.csv};
\addplot [fill=red!30,opacity=0.5,forget plot] fill between[of=case_upper and case_lower];

\addplot[name path=proposed_upper,draw=none,forget plot] table[x=a,y expr=\thisrow{Proposed}+\thisrow{Proposed-conf}, col sep=comma] {./csv/reject_stochastic_d001_q20_test2.csv};
\addplot[name path=proposed_lower,draw=none,forget plot] table[x=a,y expr=\thisrow{Proposed}-\thisrow{Proposed-conf}, col sep=comma] {./csv/reject_stochastic_d001_q20_test2.csv};
\addplot [fill=blue!30,opacity=0.5,forget plot] fill between[of=proposed_upper and proposed_lower];

\addplot[ultra thick, red, dashed] table[x=a,y=Case-control, col sep=comma]{./csv/reject_stochastic_d001_q20_test2.csv};
\addplot[ultra thick, blue] table[x=a,y=Proposed, col sep=comma]{./csv/reject_stochastic_d001_q20_test2.csv};
\addplot[ultra thick, black, dotted, domain=0:1] {0.05};

\end{axis}
\end{tikzpicture}
\label{fig:synthetic_stochastic_reject_x_s0}
} \caption{Results for the stochastic decision-maker with $b=0.2$: The shaded regions in (a) denotes the average DP $\pm$ std. The shaded regions in (b), (c), and (d) denotes 95\% confidence intervals. The dotted line in (b), (c), and (d) denotes the significance level $0.05$.}
\label{fig:app_synthetic_stochastic}
\end{figure*}

\medskip
\noindent
\textbf{Discussion \:}
The high effectiveness of the stealthily biased sampling in the stochastic decision-maker can be explained as follows.
Suppose the original dataset $D$ follows a distribution $P(y, x, s)$, and the target distribution we want to attain in the sampled dataset $\bench \subseteq D$ be $Q(y, x, s)$.
Recall that $P(y, x, s) = P(y \mid x, s) P(x, s)$ and $Q(y, x, s) = Q(y \mid x, s) Q(x, s)$.
In the deterministic case, the decision criteria is given by \eqref{eq:simple_decision} which indicates $P(y \mid x, s) = Q(y \mid x, s)$.
Hence, to modify the distribution $P(y, x, s)$ to the target distribution $Q(y, x, s)$ through sampling, we need to modify $Q(x, s)$ from $P(x, s)$.
Thus, the distribution change from $P(x, s)$ to $Q(x, s)$ can be detected by using the detector's data $D' \sim P$ and the two-sample test between $P(x, s)$ and $Q(x, s)$.
To fool the two-sample test and avoid the high rejection rate, we need to minimize the modification and keep $Q(x, s)$ close to $P(x, s)$ as much as possible.
This is the reason why the available ratio of positive decisions $\alpha$ is limited in the deterministic case.
If $\alpha$ is far from the true ratio of positive decisions in $D$, we need a large modification on $Q(x, s)$, which makes it difficult to pass the two-sample test.
By contrast, in the stochastic case, we have a chance to modify $Q(y | x, s)$ from $P(y | x, s)$ through sampling.
We can therefore modify the distribution $P(y, x, s)$ to $Q(y, x, s)$ while keeping $Q(x, s) \approx P(x, s)$ by using the stealthily biased sampling, which makes it easy to pass the two-sample test.
This is the reason why the stealthily biased sampling is highly effective in the stochastic case.
This fact implies that, if the decision-maker is interested in pretending to be fair, making stochastic decisions helps.

\section{Real-World Data: Full Results}
\label{app:real}

Here, we show the full results we omitted in Sections~\ref{sec:compas} and \ref{sec:adult} due to the space limitation.

\subsection{COMPAS Data: Setup}
\label{app:compas_setup}

For the first real-world data experiment, we focus on the COMPAS dataset~\cite{angwin2016machine}.\footnote{\url{https://github.com/propublica/compas-analysis}}
The COMPAS dataset contains several defendant's records obtained from the Broward County Sheriff’s Office in Florida.
Each defendant is scored his or her risk of recidivism using a software called COMPAS.
ProPublica~\cite{angwin2016machine} revealed that the COMPAS risk score is discriminative: it tends to score white defendants with low scores while scoring black defendants with high scores.

Because Florida had strong open-records laws, the entire COMPAS dataset was made public and the bias in the COMPAS risk score was revealed.
Here, we consider a virtual scenario that the decision-maker was aware of the bias in the risk score, and he wants to pretend to be fair by hiding the bias.
To attain this goal, the decision-maker discloses a subset of the COMPAS dataset as evidence that the COMPAS risk score is fair.

\medskip
\noindent
\textbf{Data \:}
We used the same data preprocessing following the analysis of ProPublica~\cite{angwin2016machine}, which results in eight features $x \in \mathbb{R}^8$ of each defendant, with race as sensitive attribute $s \in \{0 (\text{black}), 1 (\text{white})\}$, and the decision $y \in \{0 (\text{low-risk}), 1 (\text{middle/high-risk})\}$.
The preprocessed data includes $5,278$ records, which we randomly held out $1,278$ records as the referential dataset $D'$ for the detector.
From the remaining $4,000$ records $D$, we sampled $2,000$ records as $\bench$ using both the stealthily biased and case-control sampling.
To reduce the DP in the sampling, we required the sampled set to satisfy $\Pr ( y=1 \mid s=1 ) \approx \Pr ( y=1 \mid s=0 ) \approx \alpha$ for some $\alpha \in [0, 1]$.

\medskip
\noindent
\textbf{Detector \:}
The detector tries to detect the bias in the disclosed dataset $\bench$ by comparing its distribution with the referential dataset $D'$.
In the experiment, we adopted the Wasserstein distance (WD) as the detector's detection criteria.\footnote{In COMPAS and Adult experiments, we did not adopt the multi-dimensional two-sample tests because they were too weak.}
If the WD between $\bench$ and $D'$ is sufficiently large, the detector can detect the bias in $\bench$, and thus the fraud of the decision-maker is revealed.

\subsection{COMPAS Data: Results}
\label{app:compas_res}

We repeated the experiment 100 times by randomly changing the data splitting, and summarized the results.
The full results of the COMPAS data experiment is shown in \figurename~\ref{fig:appendix_compas}.
The figures now include the WDs on $\Pr(x \mid s=1)$ and $\Pr(x \mid s=0)$ addition to \figurename~\ref{fig:compas}.
As the baseline without any biased sampling, we computed DP and the WD for randomly sampled $2,000$ records from $D$, which are denoted as \emph{Baseline} in the figures.
The results indicate the success of the stealthily biased sampling.

\begin{figure*}[t]
\centering
\subfigure[Demographic parity]{
\centering
\begin{tikzpicture}
\begin{axis}[
scale=0.8,
width=0.45\linewidth,
height=120pt,
xmin=0.25,
xmax=0.75,
yticklabel style={
    /pgf/number format/fixed,
    /pgf/number format/precision=2
},
xlabel={Ratio of positive decisions $\alpha$ in sampling},
ylabel={Average DP},
legend style={at={(0.5,1.1)},anchor=south,legend columns=-1, font=\footnotesize}
]

\addplot[name path=original_upper,draw=none,forget plot] table[x=ypos,y expr=\thisrow{No-Sampling}+\thisrow{Np-Sampling-std}, col sep=comma] {./csv/compas_result_parity.csv};
\addplot[name path=original_lower,draw=none,forget plot] table[x=ypos,y expr=\thisrow{No-Sampling}-\thisrow{Np-Sampling-std}, col sep=comma] {./csv/compas_result_parity.csv};
\addplot [fill=black!30,opacity=0.5,forget plot] fill between[of=original_upper and original_lower];

\addplot[name path=case_upper,draw=none,forget plot] table[x=ypos,y expr=\thisrow{Case-control}+\thisrow{Case-control-std}, col sep=comma] {./csv/compas_result_parity.csv};
\addplot[name path=case_lower,draw=none,forget plot] table[x=ypos,y expr=\thisrow{Case-control}-\thisrow{Case-control-std}, col sep=comma] {./csv/compas_result_parity.csv};
\addplot [fill=red!30,opacity=0.5,forget plot] fill between[of=case_upper and case_lower];

\addplot[name path=proposed_upper,draw=none,forget plot] table[x=ypos,y expr=\thisrow{Proposed}+\thisrow{Proposed-std}, col sep=comma] {./csv/compas_result_parity.csv};
\addplot[name path=proposed_lower,draw=none,forget plot] table[x=ypos,y expr=\thisrow{Proposed}-\thisrow{Proposed-std}, col sep=comma] {./csv/compas_result_parity.csv};
\addplot [fill=blue!30,opacity=0.5,forget plot] fill between[of=proposed_upper and proposed_lower];

\addplot[ultra thick, black, dotted] table[x=ypos,y=No-Sampling, col sep=comma]{./csv/compas_result_parity.csv};
\addplot[ultra thick, red, dashed] table[x=ypos,y=Case-control, col sep=comma]{./csv/compas_result_parity.csv};
\addplot[ultra thick, blue] table[x=ypos,y=Proposed, col sep=comma]{./csv/compas_result_parity.csv};

\legend{Baseline, Case-control, Stealth}

\end{axis}
\end{tikzpicture}
\label{fig:compas_parity_app}
}
\subfigure[Wasserstein distance in $\Pr(x)$]{
\centering
\begin{tikzpicture}
\begin{axis}[
scale=0.8,
width=0.45\linewidth,
height=120pt,
xmin=0.25,
xmax=0.75,
xlabel={Ratio of positive decisions $\alpha$ in sampling},
ylabel={Average WD},
legend style={at={(0.6,0.97)},anchor=north}
]

\addplot[name path=original_upper,draw=none,forget plot] table[x=ypos,y expr=\thisrow{No-Sampling}+\thisrow{Np-Sampling-std}, col sep=comma] {./csv/compas_result_distance_test00.csv};
\addplot[name path=original_lower,draw=none,forget plot] table[x=ypos,y expr=\thisrow{No-Sampling}-\thisrow{Np-Sampling-std}, col sep=comma] {./csv/compas_result_distance_test00.csv};
\addplot [fill=black!30,opacity=0.5,forget plot] fill between[of=original_upper and original_lower];

\addplot[name path=case_upper,draw=none,forget plot] table[x=ypos,y expr=\thisrow{Case-control}+\thisrow{Case-control-std}, col sep=comma] {./csv/compas_result_distance_test00.csv};
\addplot[name path=case_lower,draw=none,forget plot] table[x=ypos,y expr=\thisrow{Case-control}-\thisrow{Case-control-std}, col sep=comma] {./csv/compas_result_distance_test00.csv};
\addplot [fill=red!30,opacity=0.5,forget plot] fill between[of=case_upper and case_lower];

\addplot[name path=proposed_upper,draw=none,forget plot] table[x=ypos,y expr=\thisrow{Proposed}+\thisrow{Proposed-std}, col sep=comma] {./csv/compas_result_distance_test00.csv};
\addplot[name path=proposed_lower,draw=none,forget plot] table[x=ypos,y expr=\thisrow{Proposed}-\thisrow{Proposed-std}, col sep=comma] {./csv/compas_result_distance_test00.csv};
\addplot [fill=blue!30,opacity=0.5,forget plot] fill between[of=proposed_upper and proposed_lower];

\addplot[ultra thick, black, dotted] table[x=ypos,y=No-Sampling, col sep=comma]{./csv/compas_result_distance_test00.csv};
\addplot[ultra thick, red, dashed] table[x=ypos,y=Case-control, col sep=comma]{./csv/compas_result_distance_test00.csv};
\addplot[ultra thick, blue] table[x=ypos,y=Proposed, col sep=comma]{./csv/compas_result_distance_test00.csv};

\end{axis}
\end{tikzpicture}
\label{fig:compas_distance_x_app}
}
\subfigure[Wasserstein distance in $\Pr(x|s=1)$]{
\centering
\begin{tikzpicture}
\begin{axis}[
scale=0.8,
width=0.45\linewidth,
height=120pt,
xmin=0.25,
xmax=0.75,
xlabel={Ratio of positive decisions $\alpha$ in sampling},
ylabel={Average WD},
legend style={at={(0.5,1.1)},anchor=south,legend columns=-1, font=\footnotesize}
]

\addplot[name path=original_upper,draw=none,forget plot] table[x=ypos,y expr=\thisrow{No-Sampling}+\thisrow{Np-Sampling-std}, col sep=comma] {./csv/compas_result_distance_test01.csv};
\addplot[name path=original_lower,draw=none,forget plot] table[x=ypos,y expr=\thisrow{No-Sampling}-\thisrow{Np-Sampling-std}, col sep=comma] {./csv/compas_result_distance_test01.csv};
\addplot [fill=black!30,opacity=0.5,forget plot] fill between[of=original_upper and original_lower];

\addplot[name path=case_upper,draw=none,forget plot] table[x=ypos,y expr=\thisrow{Case-control}+\thisrow{Case-control-std}, col sep=comma] {./csv/compas_result_distance_test01.csv};
\addplot[name path=case_lower,draw=none,forget plot] table[x=ypos,y expr=\thisrow{Case-control}-\thisrow{Case-control-std}, col sep=comma] {./csv/compas_result_distance_test01.csv};
\addplot [fill=red!30,opacity=0.5,forget plot] fill between[of=case_upper and case_lower];

\addplot[name path=proposed_upper,draw=none,forget plot] table[x=ypos,y expr=\thisrow{Proposed}+\thisrow{Proposed-std}, col sep=comma] {./csv/compas_result_distance_test01.csv};
\addplot[name path=proposed_lower,draw=none,forget plot] table[x=ypos,y expr=\thisrow{Proposed}-\thisrow{Proposed-std}, col sep=comma] {./csv/compas_result_distance_test01.csv};
\addplot [fill=blue!30,opacity=0.5,forget plot] fill between[of=proposed_upper and proposed_lower];

\addplot[ultra thick, black, dotted] table[x=ypos,y=No-Sampling, col sep=comma]{./csv/compas_result_distance_test01.csv};
\addplot[ultra thick, red, dashed] table[x=ypos,y=Case-control, col sep=comma]{./csv/compas_result_distance_test01.csv};
\addplot[ultra thick, blue] table[x=ypos,y=Proposed, col sep=comma]{./csv/compas_result_distance_test01.csv};

\end{axis}
\end{tikzpicture}
\label{fig:compas_distance_x_s1_app}
}
\subfigure[Wasserstein distance in $\Pr(x|s=0)$]{
\centering
\begin{tikzpicture}
\begin{axis}[
scale=0.8,
width=0.45\linewidth,
height=120pt,
xmin=0.25,
xmax=0.75,
xlabel={Ratio of positive decisions $\alpha$ in sampling},
ylabel={Average WD},
legend style={at={(0.47,0.97)},anchor=north}
]

\addplot[name path=original_upper,draw=none,forget plot] table[x=ypos,y expr=\thisrow{No-Sampling}+\thisrow{Np-Sampling-std}, col sep=comma] {./csv/compas_result_distance_test02.csv};
\addplot[name path=original_lower,draw=none,forget plot] table[x=ypos,y expr=\thisrow{No-Sampling}-\thisrow{Np-Sampling-std}, col sep=comma] {./csv/compas_result_distance_test02.csv};
\addplot [fill=black!30,opacity=0.5,forget plot] fill between[of=original_upper and original_lower];

\addplot[name path=case_upper,draw=none,forget plot] table[x=ypos,y expr=\thisrow{Case-control}+\thisrow{Case-control-std}, col sep=comma] {./csv/compas_result_distance_test02.csv};
\addplot[name path=case_lower,draw=none,forget plot] table[x=ypos,y expr=\thisrow{Case-control}-\thisrow{Case-control-std}, col sep=comma] {./csv/compas_result_distance_test02.csv};
\addplot [fill=red!30,opacity=0.5,forget plot] fill between[of=case_upper and case_lower];

\addplot[name path=proposed_upper,draw=none,forget plot] table[x=ypos,y expr=\thisrow{Proposed}+\thisrow{Proposed-std}, col sep=comma] {./csv/compas_result_distance_test02.csv};
\addplot[name path=proposed_lower,draw=none,forget plot] table[x=ypos,y expr=\thisrow{Proposed}-\thisrow{Proposed-std}, col sep=comma] {./csv/compas_result_distance_test02.csv};
\addplot [fill=blue!30,opacity=0.5,forget plot] fill between[of=proposed_upper and proposed_lower];

\addplot[ultra thick, black, dotted] table[x=ypos,y=No-Sampling, col sep=comma]{./csv/compas_result_distance_test02.csv};
\addplot[ultra thick, red, dashed] table[x=ypos,y=Case-control, col sep=comma]{./csv/compas_result_distance_test02.csv};
\addplot[ultra thick, blue] table[x=ypos,y=Proposed, col sep=comma]{./csv/compas_result_distance_test02.csv};

\end{axis}
\end{tikzpicture}
\label{fig:compas_distance_x_s0_app}
} \caption{Results for the COMPAS dataset: The shaded regions in (a) denotes the average DP $\pm$ std. The shaded regions in (b), (c), and (d) denote the average WD $\pm$ std.}
\label{fig:appendix_compas}
\end{figure*}

\subsection{Adult Data: Setup}
\label{app:adult_setup}

As the second real-world data experiment, we used the Adult dataset~\cite{Dua:2017}.
The Adult dataset contains 48,842 records with several individual's features and their labels (high-income or low-income).
The dataset is known to include gender bias: in the dataset, while 30\% of the male have high-income, only 10\% of the female have high-income.
The DP of the dataset is therefore $0.2$.
If we naively train a classifier using the dataset, the resulting classifier inherits the bias and becomes discriminative, i.e., the classifier favors to classify males as high-income.
The goal of this experiment is to show that as if the biased classifier is fair by disclosing a part of the dataset with classifier's decision.

\medskip
\noindent
\textbf{Data \& Classifier \:}
In the data preprocessing, we converted categorical features to numerical features.\footnote{We used the implementation used in \url{https://www.kaggle.com/kost13/us-income-logistic-regression/notebook}}
We randomly split 10,000 records for the training set, 20,000 records for the test set, and the remaining 18,842 records for the referential set $D'$ for the detector.
In the experiment, we first train a classifier using the training set.
As a classifier, we used logistic regression and random forest with $100$ trees.
We labeled all the records in the test set using the trained classifier and obtained the dataset $D$ with the classifier's decision.
We then sample the subset $\bench \subseteq D$ with size $|\bench| = 2,000$ using both the stealthily biased and case-control sampling.
To reduce the DP in the sampling, we required the sampled set to satisfy $\Pr ( y=1 \mid s=1 ) \approx \Pr ( y=1 \mid s=0 ) \approx \alpha$ for a predetermined ratio of positive decisions $\alpha \in [0, 1]$.

\medskip
\noindent
\textbf{Detector \:}
We adopted the same detector as the COMPAS data experiment, who refers to the WD as the bias detection metric.

\subsection{Adult Data: Results}
\label{app:adult_res}

We repeated the experiment 100 times by randomly changing the data splitting, and summarized the results.
As the baseline, we computed the DP and the WD for randomly sampled $2,000$ sampled records from $D$, which are denoted as \emph{Baseline} in the figures.

\medskip
\noindent
\textbf{Logistic Regression \:}
We show the full results of the Adult data experiment in \figurename~\ref{fig:appendix_adult}.
The figures now include the WDs on $\Pr(x \mid s=1)$ and $\Pr(x \mid s=0)$ addition to \figurename~\ref{fig:adult}.
The results indicate the success of the stealthily biased sampling.

\medskip
\noindent
\textbf{Random Forest \:}
We show the results for the random forest on the Adult data experiment in \figurename~\ref{fig:appendix_adult_forest}.
The results were almost the same as logistic regression.

\begin{figure*}[t]
\centering
\subfigure[Demographic parity]{
\centering
\begin{tikzpicture}
\begin{axis}[
scale=0.8,
width=0.45\linewidth,
height=120pt,
xmin=0.05,
xmax=0.5,
yticklabel style={
    /pgf/number format/fixed,
    /pgf/number format/precision=2
},
xlabel={Ratio of positive decisions $\alpha$ in sampling},
ylabel={Average DP},
legend style={at={(0.5,1.1)},anchor=south,legend columns=-1, font=\footnotesize}
]

\addplot[name path=original_upper,draw=none,forget plot] table[x=ypos,y expr=\thisrow{No-Sampling}+\thisrow{Np-Sampling-std}, col sep=comma] {./csv/result_LogReg_parity.csv};
\addplot[name path=original_lower,draw=none,forget plot] table[x=ypos,y expr=\thisrow{No-Sampling}-\thisrow{Np-Sampling-std}, col sep=comma] {./csv/result_LogReg_parity.csv};
\addplot [fill=black!30,opacity=0.5,forget plot] fill between[of=original_upper and original_lower];

\addplot[name path=case_upper,draw=none,forget plot] table[x=ypos,y expr=\thisrow{Case-control}+\thisrow{Case-control-std}, col sep=comma] {./csv/result_LogReg_parity.csv};
\addplot[name path=case_lower,draw=none,forget plot] table[x=ypos,y expr=\thisrow{Case-control}-\thisrow{Case-control-std}, col sep=comma] {./csv/result_LogReg_parity.csv};
\addplot [fill=red!30,opacity=0.5,forget plot] fill between[of=case_upper and case_lower];

\addplot[name path=proposed_upper,draw=none,forget plot] table[x=ypos,y expr=\thisrow{Proposed}+\thisrow{Proposed-std}, col sep=comma] {./csv/result_LogReg_parity.csv};
\addplot[name path=proposed_lower,draw=none,forget plot] table[x=ypos,y expr=\thisrow{Proposed}-\thisrow{Proposed-std}, col sep=comma] {./csv/result_LogReg_parity.csv};
\addplot [fill=blue!30,opacity=0.5,forget plot] fill between[of=proposed_upper and proposed_lower];

\addplot[ultra thick, black, dotted] table[x=ypos,y=No-Sampling, col sep=comma]{./csv/result_LogReg_parity.csv};
\addplot[ultra thick, red, dashed] table[x=ypos,y=Case-control, col sep=comma]{./csv/result_LogReg_parity.csv};
\addplot[ultra thick, blue] table[x=ypos,y=Proposed, col sep=comma]{./csv/result_LogReg_parity.csv};

\legend{Baseline, Case-control, Stealth}

\end{axis}
\end{tikzpicture}
\label{fig:logreg_parity_app}
}
\subfigure[Wasserstein Distance in $\Pr(x)$]{
\centering
\begin{tikzpicture}
\begin{axis}[
scale=0.8,
width=0.45\linewidth,
height=120pt,
xmin=0.05,
xmax=0.5,
xlabel={Ratio of positive decisions $\alpha$ in sampling},
ylabel={Average WD},
legend style={at={(0.37,0.97)},anchor=north}
]

\addplot[name path=original_upper,draw=none,forget plot] table[x=ypos,y expr=\thisrow{No-Sampling}+\thisrow{Np-Sampling-std}, col sep=comma] {./csv/result_LogReg_distance_test00.csv};
\addplot[name path=original_lower,draw=none,forget plot] table[x=ypos,y expr=\thisrow{No-Sampling}-\thisrow{Np-Sampling-std}, col sep=comma] {./csv/result_LogReg_distance_test00.csv};
\addplot [fill=black!30,opacity=0.5,forget plot] fill between[of=original_upper and original_lower];

\addplot[name path=case_upper,draw=none,forget plot] table[x=ypos,y expr=\thisrow{Case-control}+\thisrow{Case-control-std}, col sep=comma] {./csv/result_LogReg_distance_test00.csv};
\addplot[name path=case_lower,draw=none,forget plot] table[x=ypos,y expr=\thisrow{Case-control}-\thisrow{Case-control-std}, col sep=comma] {./csv/result_LogReg_distance_test00.csv};
\addplot [fill=red!30,opacity=0.5,forget plot] fill between[of=case_upper and case_lower];

\addplot[name path=proposed_upper,draw=none,forget plot] table[x=ypos,y expr=\thisrow{Proposed}+\thisrow{Proposed-std}, col sep=comma] {./csv/result_LogReg_distance_test00.csv};
\addplot[name path=proposed_lower,draw=none,forget plot] table[x=ypos,y expr=\thisrow{Proposed}-\thisrow{Proposed-std}, col sep=comma] {./csv/result_LogReg_distance_test00.csv};
\addplot [fill=blue!30,opacity=0.5,forget plot] fill between[of=proposed_upper and proposed_lower];

\addplot[ultra thick, black, dotted] table[x=ypos,y=No-Sampling, col sep=comma]{./csv/result_LogReg_distance_test00.csv};
\addplot[ultra thick, red, dashed] table[x=ypos,y=Case-control, col sep=comma]{./csv/result_LogReg_distance_test00.csv};
\addplot[ultra thick, blue] table[x=ypos,y=Proposed, col sep=comma]{./csv/result_LogReg_distance_test00.csv};

\end{axis}
\end{tikzpicture}
\label{fig:logreg_distance_x_app}
}
\subfigure[Wasserstein Distance in $\Pr(x|s=1)$]{
\centering
\begin{tikzpicture}
\begin{axis}[
scale=0.8,
width=0.45\linewidth,
height=120pt,
xmin=0.05,
xmax=0.5,
xlabel={Ratio of positive decisions $\alpha$ in sampling},
ylabel={Average WD},
legend style={at={(0.47,0.97)},anchor=north}
]

\addplot[name path=original_upper,draw=none,forget plot] table[x=ypos,y expr=\thisrow{No-Sampling}+\thisrow{Np-Sampling-std}, col sep=comma] {./csv/result_LogReg_distance_test01.csv};
\addplot[name path=original_lower,draw=none,forget plot] table[x=ypos,y expr=\thisrow{No-Sampling}-\thisrow{Np-Sampling-std}, col sep=comma] {./csv/result_LogReg_distance_test01.csv};
\addplot [fill=black!30,opacity=0.5,forget plot] fill between[of=original_upper and original_lower];

\addplot[name path=case_upper,draw=none,forget plot] table[x=ypos,y expr=\thisrow{Case-control}+\thisrow{Case-control-std}, col sep=comma] {./csv/result_LogReg_distance_test01.csv};
\addplot[name path=case_lower,draw=none,forget plot] table[x=ypos,y expr=\thisrow{Case-control}-\thisrow{Case-control-std}, col sep=comma] {./csv/result_LogReg_distance_test01.csv};
\addplot [fill=red!30,opacity=0.5,forget plot] fill between[of=case_upper and case_lower];

\addplot[name path=proposed_upper,draw=none,forget plot] table[x=ypos,y expr=\thisrow{Proposed}+\thisrow{Proposed-std}, col sep=comma] {./csv/result_LogReg_distance_test01.csv};
\addplot[name path=proposed_lower,draw=none,forget plot] table[x=ypos,y expr=\thisrow{Proposed}-\thisrow{Proposed-std}, col sep=comma] {./csv/result_LogReg_distance_test01.csv};
\addplot [fill=blue!30,opacity=0.5,forget plot] fill between[of=proposed_upper and proposed_lower];

\addplot[ultra thick, black, dotted] table[x=ypos,y=No-Sampling, col sep=comma]{./csv/result_LogReg_distance_test01.csv};
\addplot[ultra thick, red, dashed] table[x=ypos,y=Case-control, col sep=comma]{./csv/result_LogReg_distance_test01.csv};
\addplot[ultra thick, blue] table[x=ypos,y=Proposed, col sep=comma]{./csv/result_LogReg_distance_test01.csv};

\end{axis}
\end{tikzpicture}
\label{fig:logreg_distance_x_s1_app}
}
\subfigure[Wasserstein Distance in $\Pr(x|s=0)$]{
\centering
\begin{tikzpicture}
\begin{axis}[
scale=0.8,
width=0.45\linewidth,
height=120pt,
xmin=0.05,
xmax=0.5,
xlabel={Ratio of positive decisions $\alpha$ in sampling},
ylabel={Average WD},
legend style={at={(0.47,0.97)},anchor=north}
]

\addplot[name path=original_upper,draw=none,forget plot] table[x=ypos,y expr=\thisrow{No-Sampling}+\thisrow{Np-Sampling-std}, col sep=comma] {./csv/result_LogReg_distance_test02.csv};
\addplot[name path=original_lower,draw=none,forget plot] table[x=ypos,y expr=\thisrow{No-Sampling}-\thisrow{Np-Sampling-std}, col sep=comma] {./csv/result_LogReg_distance_test02.csv};
\addplot [fill=black!30,opacity=0.5,forget plot] fill between[of=original_upper and original_lower];

\addplot[name path=case_upper,draw=none,forget plot] table[x=ypos,y expr=\thisrow{Case-control}+\thisrow{Case-control-std}, col sep=comma] {./csv/result_LogReg_distance_test02.csv};
\addplot[name path=case_lower,draw=none,forget plot] table[x=ypos,y expr=\thisrow{Case-control}-\thisrow{Case-control-std}, col sep=comma] {./csv/result_LogReg_distance_test02.csv};
\addplot [fill=red!30,opacity=0.5,forget plot] fill between[of=case_upper and case_lower];

\addplot[name path=proposed_upper,draw=none,forget plot] table[x=ypos,y expr=\thisrow{Proposed}+\thisrow{Proposed-std}, col sep=comma] {./csv/result_LogReg_distance_test02.csv};
\addplot[name path=proposed_lower,draw=none,forget plot] table[x=ypos,y expr=\thisrow{Proposed}-\thisrow{Proposed-std}, col sep=comma] {./csv/result_LogReg_distance_test02.csv};
\addplot [fill=blue!30,opacity=0.5,forget plot] fill between[of=proposed_upper and proposed_lower];

\addplot[ultra thick, black, dotted] table[x=ypos,y=No-Sampling, col sep=comma]{./csv/result_LogReg_distance_test02.csv};
\addplot[ultra thick, red, dashed] table[x=ypos,y=Case-control, col sep=comma]{./csv/result_LogReg_distance_test02.csv};
\addplot[ultra thick, blue] table[x=ypos,y=Proposed, col sep=comma]{./csv/result_LogReg_distance_test02.csv};

\end{axis}
\end{tikzpicture}
\label{fig:logreg_distance_x_s0_app}
} \caption{[Logistic Regression] Results for the Adult dataset: The shaded regions in (a) denotes the average DP $\pm$ std. The shaded regions in (b), (c), and (d) denote the average WD $\pm$ std.}
\label{fig:appendix_adult}
\vspace{24pt}
\centering
\subfigure[Demographic parity]{
\centering
\begin{tikzpicture}
\begin{axis}[
scale=0.8,
width=0.45\linewidth,
height=120pt,
xmin=0.05,
xmax=0.5,
yticklabel style={
    /pgf/number format/fixed,
    /pgf/number format/precision=2
},
xlabel={Ratio of positive decisions $\alpha$ in sampling},
ylabel={Average DP},
legend style={at={(0.5,1.1)},anchor=south,legend columns=-1, font=\footnotesize}
]

\addplot[name path=original_upper,draw=none,forget plot] table[x=ypos,y expr=\thisrow{No-Sampling}+\thisrow{Np-Sampling-std}, col sep=comma] {./csv/result_Forest_parity.csv};
\addplot[name path=original_lower,draw=none,forget plot] table[x=ypos,y expr=\thisrow{No-Sampling}-\thisrow{Np-Sampling-std}, col sep=comma] {./csv/result_Forest_parity.csv};
\addplot [fill=black!30,opacity=0.5,forget plot] fill between[of=original_upper and original_lower];

\addplot[name path=case_upper,draw=none,forget plot] table[x=ypos,y expr=\thisrow{Case-control}+\thisrow{Case-control-std}, col sep=comma] {./csv/result_Forest_parity.csv};
\addplot[name path=case_lower,draw=none,forget plot] table[x=ypos,y expr=\thisrow{Case-control}-\thisrow{Case-control-std}, col sep=comma] {./csv/result_Forest_parity.csv};
\addplot [fill=red!30,opacity=0.5,forget plot] fill between[of=case_upper and case_lower];

\addplot[name path=proposed_upper,draw=none,forget plot] table[x=ypos,y expr=\thisrow{Proposed}+\thisrow{Proposed-std}, col sep=comma] {./csv/result_Forest_parity.csv};
\addplot[name path=proposed_lower,draw=none,forget plot] table[x=ypos,y expr=\thisrow{Proposed}-\thisrow{Proposed-std}, col sep=comma] {./csv/result_Forest_parity.csv};
\addplot [fill=blue!30,opacity=0.5,forget plot] fill between[of=proposed_upper and proposed_lower];

\addplot[ultra thick, black, dotted] table[x=ypos,y=No-Sampling, col sep=comma]{./csv/result_Forest_parity.csv};
\addplot[ultra thick, red, dashed] table[x=ypos,y=Case-control, col sep=comma]{./csv/result_Forest_parity.csv};
\addplot[ultra thick, blue] table[x=ypos,y=Proposed, col sep=comma]{./csv/result_Forest_parity.csv};

\legend{Baseline, Case-control, Stealth}

\end{axis}
\end{tikzpicture}
\label{fig:Forest_parity_app}
}
\subfigure[Wasserstein Distance in $\Pr(x)$]{
\centering
\begin{tikzpicture}
\begin{axis}[
scale=0.8,
width=0.45\linewidth,
height=120pt,
xmin=0.05,
xmax=0.5,
xlabel={Ratio of positive decisions $\alpha$ in sampling},
ylabel={Average WD},
legend style={at={(0.37,0.97)},anchor=north}
]

\addplot[name path=original_upper,draw=none,forget plot] table[x=ypos,y expr=\thisrow{No-Sampling}+\thisrow{Np-Sampling-std}, col sep=comma] {./csv/result_Forest_distance_test00.csv};
\addplot[name path=original_lower,draw=none,forget plot] table[x=ypos,y expr=\thisrow{No-Sampling}-\thisrow{Np-Sampling-std}, col sep=comma] {./csv/result_Forest_distance_test00.csv};
\addplot [fill=black!30,opacity=0.5,forget plot] fill between[of=original_upper and original_lower];

\addplot[name path=case_upper,draw=none,forget plot] table[x=ypos,y expr=\thisrow{Case-control}+\thisrow{Case-control-std}, col sep=comma] {./csv/result_Forest_distance_test00.csv};
\addplot[name path=case_lower,draw=none,forget plot] table[x=ypos,y expr=\thisrow{Case-control}-\thisrow{Case-control-std}, col sep=comma] {./csv/result_Forest_distance_test00.csv};
\addplot [fill=red!30,opacity=0.5,forget plot] fill between[of=case_upper and case_lower];

\addplot[name path=proposed_upper,draw=none,forget plot] table[x=ypos,y expr=\thisrow{Proposed}+\thisrow{Proposed-std}, col sep=comma] {./csv/result_Forest_distance_test00.csv};
\addplot[name path=proposed_lower,draw=none,forget plot] table[x=ypos,y expr=\thisrow{Proposed}-\thisrow{Proposed-std}, col sep=comma] {./csv/result_Forest_distance_test00.csv};
\addplot [fill=blue!30,opacity=0.5,forget plot] fill between[of=proposed_upper and proposed_lower];

\addplot[ultra thick, black, dotted] table[x=ypos,y=No-Sampling, col sep=comma]{./csv/result_Forest_distance_test00.csv};
\addplot[ultra thick, red, dashed] table[x=ypos,y=Case-control, col sep=comma]{./csv/result_Forest_distance_test00.csv};
\addplot[ultra thick, blue] table[x=ypos,y=Proposed, col sep=comma]{./csv/result_Forest_distance_test00.csv};

\end{axis}
\end{tikzpicture}
\label{fig:Forest_distance_x_app}
}
\subfigure[Wasserstein Distance in $\Pr(x|s=1)$]{
\centering
\begin{tikzpicture}
\begin{axis}[
scale=0.8,
width=0.45\linewidth,
height=120pt,
xmin=0.05,
xmax=0.5,
xlabel={Ratio of positive decisions $\alpha$ in sampling},
ylabel={Average WD},
legend style={at={(0.5,1.1)},anchor=south,legend columns=-1, font=\footnotesize}
]

\addplot[name path=original_upper,draw=none,forget plot] table[x=ypos,y expr=\thisrow{No-Sampling}+\thisrow{Np-Sampling-std}, col sep=comma] {./csv/result_Forest_distance_test01.csv};
\addplot[name path=original_lower,draw=none,forget plot] table[x=ypos,y expr=\thisrow{No-Sampling}-\thisrow{Np-Sampling-std}, col sep=comma] {./csv/result_Forest_distance_test01.csv};
\addplot [fill=black!30,opacity=0.5,forget plot] fill between[of=original_upper and original_lower];

\addplot[name path=case_upper,draw=none,forget plot] table[x=ypos,y expr=\thisrow{Case-control}+\thisrow{Case-control-std}, col sep=comma] {./csv/result_Forest_distance_test01.csv};
\addplot[name path=case_lower,draw=none,forget plot] table[x=ypos,y expr=\thisrow{Case-control}-\thisrow{Case-control-std}, col sep=comma] {./csv/result_Forest_distance_test01.csv};
\addplot [fill=red!30,opacity=0.5,forget plot] fill between[of=case_upper and case_lower];

\addplot[name path=proposed_upper,draw=none,forget plot] table[x=ypos,y expr=\thisrow{Proposed}+\thisrow{Proposed-std}, col sep=comma] {./csv/result_Forest_distance_test01.csv};
\addplot[name path=proposed_lower,draw=none,forget plot] table[x=ypos,y expr=\thisrow{Proposed}-\thisrow{Proposed-std}, col sep=comma] {./csv/result_Forest_distance_test01.csv};
\addplot [fill=blue!30,opacity=0.5,forget plot] fill between[of=proposed_upper and proposed_lower];

\addplot[ultra thick, black, dotted] table[x=ypos,y=No-Sampling, col sep=comma]{./csv/result_Forest_distance_test01.csv};
\addplot[ultra thick, red, dashed] table[x=ypos,y=Case-control, col sep=comma]{./csv/result_Forest_distance_test01.csv};
\addplot[ultra thick, blue] table[x=ypos,y=Proposed, col sep=comma]{./csv/result_Forest_distance_test01.csv};

\end{axis}
\end{tikzpicture}
\label{fig:Forest_distance_x_s1_app}
}
\subfigure[Wasserstein Distance in $\Pr(x|s=0)$]{
\centering
\begin{tikzpicture}
\begin{axis}[
scale=0.8,
width=0.45\linewidth,
height=120pt,
xmin=0.05,
xmax=0.5,
xlabel={Ratio of positive decisions $\alpha$ in sampling},
ylabel={Average WD},
legend style={at={(0.47,0.97)},anchor=north}
]

\addplot[name path=original_upper,draw=none,forget plot] table[x=ypos,y expr=\thisrow{No-Sampling}+\thisrow{Np-Sampling-std}, col sep=comma] {./csv/result_Forest_distance_test02.csv};
\addplot[name path=original_lower,draw=none,forget plot] table[x=ypos,y expr=\thisrow{No-Sampling}-\thisrow{Np-Sampling-std}, col sep=comma] {./csv/result_Forest_distance_test02.csv};
\addplot [fill=black!30,opacity=0.5,forget plot] fill between[of=original_upper and original_lower];

\addplot[name path=case_upper,draw=none,forget plot] table[x=ypos,y expr=\thisrow{Case-control}+\thisrow{Case-control-std}, col sep=comma] {./csv/result_Forest_distance_test02.csv};
\addplot[name path=case_lower,draw=none,forget plot] table[x=ypos,y expr=\thisrow{Case-control}-\thisrow{Case-control-std}, col sep=comma] {./csv/result_Forest_distance_test02.csv};
\addplot [fill=red!30,opacity=0.5,forget plot] fill between[of=case_upper and case_lower];

\addplot[name path=proposed_upper,draw=none,forget plot] table[x=ypos,y expr=\thisrow{Proposed}+\thisrow{Proposed-std}, col sep=comma] {./csv/result_Forest_distance_test02.csv};
\addplot[name path=proposed_lower,draw=none,forget plot] table[x=ypos,y expr=\thisrow{Proposed}-\thisrow{Proposed-std}, col sep=comma] {./csv/result_Forest_distance_test02.csv};
\addplot [fill=blue!30,opacity=0.5,forget plot] fill between[of=proposed_upper and proposed_lower];

\addplot[ultra thick, black, dotted] table[x=ypos,y=No-Sampling, col sep=comma]{./csv/result_Forest_distance_test02.csv};
\addplot[ultra thick, red, dashed] table[x=ypos,y=Case-control, col sep=comma]{./csv/result_Forest_distance_test02.csv};
\addplot[ultra thick, blue] table[x=ypos,y=Proposed, col sep=comma]{./csv/result_Forest_distance_test02.csv};

\end{axis}
\end{tikzpicture}
\label{fig:Forest_distance_x_s0_app}
} \caption{[Random Forest] Results for the Adult dataset: The shaded regions in (a) denotes the average DP $\pm$ std. The shaded regions in (b), (c), and (d) denote the average WD $\pm$ std.}
\label{fig:appendix_adult_forest}
\end{figure*}

\end{document}